%% file: manuscript.tex
\documentclass[12pt]{article}
\usepackage[T1]{fontenc}
\usepackage{array,longtable,multirow}
\usepackage{xcolor}
\usepackage{fancyhdr}
\usepackage{float}
\usepackage{graphicx}
\usepackage{indentfirst}

\usepackage[american]{babel}
\usepackage{csquotes}
\usepackage[backend=biber,citestyle=authoryear,sortcites=true,natbib,sorting=nyt,style=apa]{biblatex}
\DeclareLanguageMapping{american}{american-apa}
\usepackage{hyperref}
\usepackage{enumitem}
\usepackage[normalem]{ulem}
\newcommand{\stkout}[1]{\ifmmode\text{\sout{\ensuremath{#1}}}\else\sout{#1}\fi}
\usepackage{algorithm}
\usepackage{algorithmic}
\usepackage[
  top=2.5cm, bottom=2.5cm,
  left=2.3cm, right=2.3cm,
  headsep=3mm,
  headheight=15pt]{geometry}
\usepackage{newpxtext}
\usepackage{amsthm, amsmath, amssymb}
\usepackage{cases}
\usepackage{subcaption}
\usepackage[affil-it]{authblk}
\usepackage{textcomp}
\usepackage{booktabs}
\usepackage{multirow}

\newcommand\cP{{\cal{P}}}

\newcommand\bfx{\mathbf{x}}

\newcommand\M{\mathbf{M}}

\newcommand\rank{\mathrm{rank}}

\newcommand\PI{\mathrm{PI}}

\newcommand\datasim{Fluo-N2DH-SIM+} 
\newcommand\datagowt{Fluo-N2DH-GOWT1}
\newcommand\SEG{\mathrm{SEG}}
\newcommand\TRA{\mathrm{TRA}}
\newcommand\OPCTB{$\text{OP}_{\text{CTB}}$}

\DeclareMathOperator*{\argmax}{arg\,max}

\newtheorem{remark}{Remark}
\newtheorem{proposition}{Proposition}
\newlength{\twosubht}
\newsavebox{\twosubbox}
\usepackage{pgf}
\usepackage{import}
\title{Appearance-free Tripartite Matching for Multiple Object Tracking}
\author[1]{Lijun Wang%
  \thanks{Email: \texttt{ljwang@link.cuhk.edu.hk}}
}
\affil[1]{Department of Statistics, The Chinese University of Hong Kong, Hong Kong SAR, China}

\author[2]{Yanting Zhu}
\author[2]{Jue Shi}
\affil[2]{Department of Physics, Hong Kong Baptist University, Hong Kong SAR, China}
\author[1]{Xiaodan Fan%
  \thanks{Email: \texttt{xfan@cuhk.edu.hk}; Corresponding author}
}

\date{\today}
\begin{document}
\maketitle

\begin{abstract}
Multiple Object Tracking (MOT) detects the trajectories of multiple objects given an input video. It has become more and more important for various research and industry areas, such as cell tracking for biomedical research and human tracking in video surveillance. Most existing algorithms depend on the uniqueness of the object's appearance, and the dominating bipartite matching scheme ignores the speed smoothness. Although several methods have incorporated the velocity smoothness for tracking, they either fail to pursue global smooth velocity or are often trapped in local optimums. We focus on the general MOT problem regardless of the appearance and propose an appearance-free tripartite matching to avoid the irregular velocity problem of the bipartite matching. The tripartite matching is formulated as maximizing the likelihood of the state vectors constituted of the position and velocity of objects, which results in a chain-dependent structure. We resort to the dynamic programming algorithm to find such a maximum likelihood estimate. To overcome the high computational cost induced by the vast search space of dynamic programming when many objects are to be tracked, we decompose the space by the number of disappearing objects and propose a reduced-space approach by truncating the decomposition. Extensive simulations have shown the superiority and efficiency of our proposed method, and the comparisons with top methods on Cell Tracking Challenge also demonstrate our competence. We also applied our method to track the motion of natural killer cells around tumor cells in a cancer study.\footnote{The source code is available on \url{https://github.com/szcf-weiya/TriMatchMOT}}
\end{abstract}

\section{Introduction}\label{sec:intro}
Natural Killer (NK) cells are innate immune cells that control certain microbial infections and tumors \parencite{cerwenkaNaturalKillerCells2001}.
The background of Figure \ref{fig:realall}, i.e., the part excluding the orange and blue curves and the red rectangle box, is a frame from a cell video, where the NK cells are the brightest, roughly round, and can move freely, while the cancer cells are dimmer, flat and still. The NK cells keep bumping into the cancer cell, and eventually, the cancer cell bursts. The goal for this cell video is to track each NK cell along with the time frame. The tracking results can be used for various downstream analyses, such as the motility properties of NK cells, the associated chemotaxis studies \parencite{ferlazzoNaturalKillerCell2012}, and the assessment of crosstalk effects with other immune cells \parencite{hariziReciprocalCrosstalkDendritic2013}.

\subsection{General Multiple Object Tracking}

\begin{figure}[H]
  \centering
  \includegraphics[width=\textwidth]{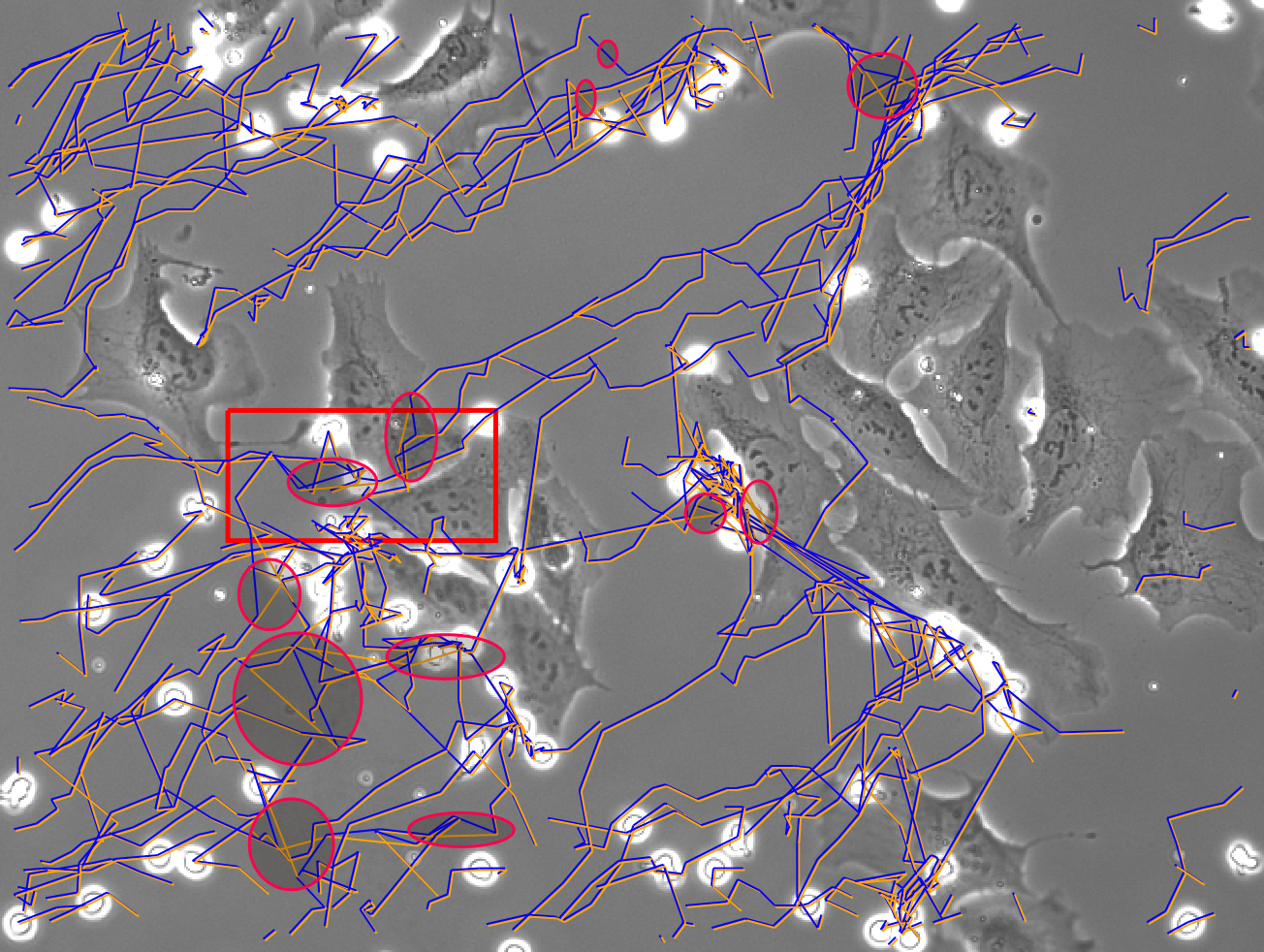}
  \caption{The background is set as one frame of the video, where the brightest small near-circles are the cells to be tracked, and other dimmer irregular contours are the still cancer cell. The orange curves represent our proposed tripartite matched trajectories, while the blue curves denote the bipartite matched paths. The red ellipses annotate the difference obtained by these two methods, and the red rectangle region will be investigated as case studies in Section \ref{sec:real-cell-video}.}
  \label{fig:realall}
\end{figure}

Tracking the NK cells is a typical task of Multiple Object Tracking (MOT), which becomes more and more popular in numerous scientific and industrious areas, such as human tracking in video surveillance or sports analysis \parencite{camplaniMultipleHumanTracking2016}, and cell tracking in cancer research or single-cell studies \parencite{maskaBenchmarkComparisonCell2014}. MOT aims to reconstruct the moving paths of multiple objects from a video, which is constituted by a series of consecutive images, where the coordinates of objects are determined by extracting their features from the images, known as object detection. Nowadays, this is a classical but still challenging problem. There are some ongoing public challenges, e.g., Cell Tracking Challenge (\url{http://celltrackingchallenge.net/}) and Multiple Human Tracking (\url{https://motchallenge.net/}), both of which provide some public datasets and attract researchers to develop their methods and compete. Extensive research in multiple object tracking has resulted in versatile and powerful algorithms. We can group these algorithms by numerous criteria \parencite{luoMultipleObjectTracking2014}. For instance, some algorithms process the video frame-by-frame, and the trajectories are estimated based only on the historical frames; this is known as online methods. In contrast, the offline methods require all frames in advance, and analyze them jointly to output the final trajectories.
Despite the huge variety of methods in the literature, many MOT algorithms take advantage of the unique appearance of each object,
such as the template matching \parencite{xiangLearningTrackOnline2015},
the level set method \parencite{yangCellSegmentationTracking2005},
the representations by deep neural networks \parencite{ciaparroneDeepLearningVideo2020},
and the overlap-based association strategies \parencite{bochinskiHighSpeedTrackingbydetectionUsing2017}. However, in some tracking tasks, we cannot expect much information from the appearance when all objects look similar, such as the roughly identical-sized round-shape Natural Killer (NK) cells in Figure \ref{fig:realall}.
It is reasonable and natural to assume that the shape and size of all objects are the same, then we resort to the motion of objects and propose the velocity model, which also refers to the tripartite model. Without a specified appearance, we can investigate and quantify the performance of the pure motion model for general objects by treating them as particles if their sizes are similar and relatively small compared to the whole tracking region.

Generally, most MOT algorithms share the following two stages,
\begin{itemize}
  \item Detection (Segmentation) stage: identify objects from each frame of the input video;
  \item Association stage: associate the objects by the detected appearance and the motion predictions.
\end{itemize}
These two stages can be conducted in different manners.

\subsection{Three Different Manners}

One popular manner is to intertwine these two stages, i.e., associating after detecting the current frame and then detecting the next frame, again followed by associating. The stochastic filter methods, in particular the Kalman filter \parencite{reidAlgorithmTrackingMultiple1979}, are the representative approaches. In general, they use a series of measurements observed over time, containing statistical noise and other inaccuracies, and produce estimates of unknown variables by estimating the posterior distribution of the variables for each time frame. For the tracking tasks, they treat the detections as the noisy measurement and assume the real states (such as coordinates) of the objects are the unknown variables.
In MOT, each object has its own state and measurement, but we do not know the correspondence between the states and measurements, so the association stage needs to be performed to determine each state's measurement, such as the feature matching step in \textcite{liMultipleObjectTracking2010}.

\begin{figure}[H]
\sbox\twosubbox{%
  \resizebox{\dimexpr\textwidth-1em}{!}{%
    \includegraphics[page=1,height=8cm]{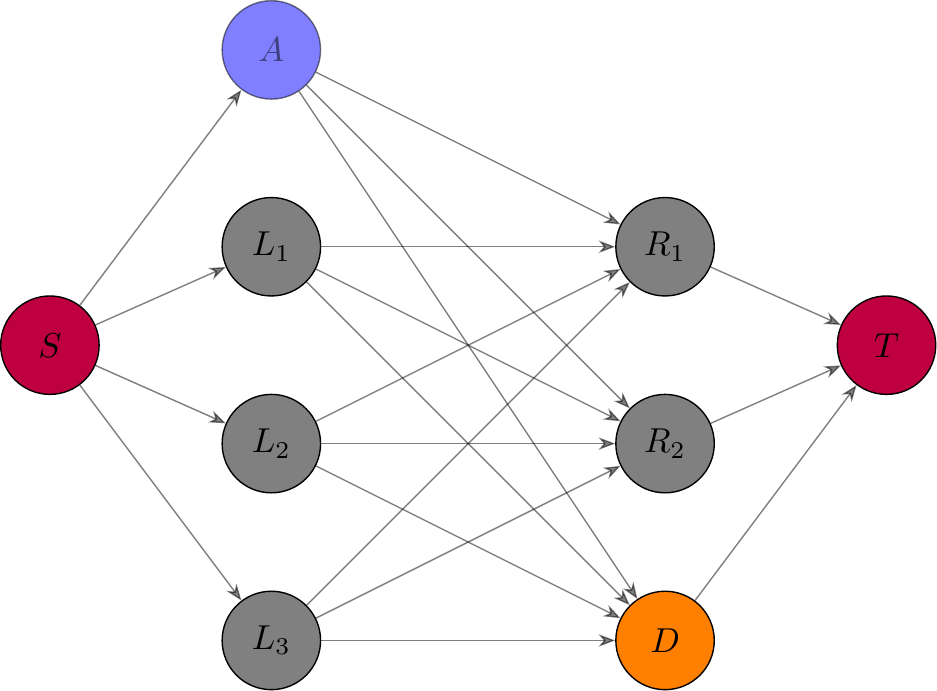}%
    \includegraphics[page=2,height=8cm]{figs/mincost2.pdf}%
  }%
}
\setlength{\twosubht}{\ht\twosubbox}
\centering
\subcaptionbox{Bipartite graph matching.\label{fig:mincost1}}{%
  \includegraphics[page=1,height=\twosubht]{figs/mincost2.pdf}%
}\,
\subcaptionbox{Graph-based global association.\label{fig:mincost2}}{%
  \includegraphics[page=2,height=\twosubht]{figs/mincost2.pdf}%
}
\caption{Two popular association approaches. (\emph{a}) The bipartite graph matching adapted from \textcite{padfieldCoupledMinimumcostFlow2011}, where the vertices labeled with $L$ represent the objects in the previous frame, and the vertices denoted by $R$ on the right denote the objects on the current frame. The appearing vertex $A$ on the left allows objects in the current frame to be newcomers, and the disappearing vertex $D$ on the right allows objects in the previous frame to leave out. (\emph{b}) The graph-based global association adapted from \textcite{zhangGlobalDataAssociation2008}, illustrated by a toy example with 3 frames and 9 observations, where the vertices at the same column mean that they are in the same frame. Each flow from source $S$ to target $T$ represents the path of a particular object, such as the thick arrowed path $S\rightarrow O_2\rightarrow O_5\rightarrow T$.}
\label{fig:mincost}
\end{figure}

Another widely used manner is to perform the detection and association stages separately, detecting all frames at once and then associating the detected objects across adjacent frames. In this case, some algorithms formulate the tracking task as an assignment problem, such as bipartite graph matching \parencite{padfieldCoupledMinimumcostFlow2011} and graph-based global data association \parencite{zhangGlobalDataAssociation2008}, both of which can be solved by a minimum-cost flow network.
\textcite{ulmanObjectiveComparisonCelltracking2017} summarizes 21 participating algorithms in the Cell Tracking Challenge, where 7 algorithms use the distance-based bipartite matching, 6 algorithms adopt the graph-based global linking methodology and \textcite{magnussonGlobalLinkingCell2015} as one of them performed extraordinarily well, which ranked among the top-three algorithms for all competing data sets. As shown in Figure \ref{fig:mincost}, apart from the source $S$ and terminal $T$, the nodes in each column represent the observations in one frame. The edges between nodes from different columns have some cost defined by particular distance metrics, such as Euclidean distance, or probabilities based on some parametric models. If we send a unit of flow over an edge, the corresponding cost would be incurred, and our goal is to send some units of flow from source $S$ to terminal $T$ in such a way that the total cost is minimized.
The bipartite matching sequentially processes only two adjacent frames, and each flow from $S$ to $T$ in Figure \ref{fig:mincost1} means a fragment of one trajectory. The appearing node $A$ serves as an internal node for bridging the source $S$ and the right node $R_j$; similarly, the disappearing node $D$ connects the terminal $T$ and the left node $L_i$, but Figure \ref{fig:mincost2} skips them by allowing the direct links from $S$ or to $T$. Moreover, the global linking puts all frames into the graph, and each flow represents a complete trajectory of a particular object.

In addition to the above two manners, some algorithms would repeatedly perform these two stages, i.e., employing the detection stage or the association stage multiple times using different techniques. \textcite{jaqamanRobustSingleparticleTracking2008}'s LAP method embedded in the popular cell image analysis software CellProfiler \parencite{carpenterCellProfilerImageAnalysis2006} is one of them. The algorithm first links detections into tracklets (fragments of the trajectory) and then links the tracklets into longer tracks by solving a combinational optimization problem. The tracklets are created and linked into tracks by solving two different Linear Assignment Problems (LAPs), where the first LAP can be viewed as distance-based bipartite matching in that it links objects for every two adjacent frames. The second LAP seems like post-processing, which allows tracklets to bridge (linking the end of a tracklet to the start of another tracklet), merge (connecting the end of a tracklet to the middle point of another tracklet), or split (joining the start of a tracklet with the middle point of another tracklet).

There are undoubtedly other methods that fall outside of the above three manners and even do not involve these two stages due to countless different applications, such as the detection-free tracking mentioned in \textcite{luoMultipleObjectTracking2014}.

With the rapid advance of deep learning, the second manner becomes more and more popular. Various deep neural networks have been applied in the segmentation stage, such as deep Convolutional Neural Networks (CNN) for fluorescently labeled cells \parencite{sadanandanAutomatedTrainingDeep2017}, Mask R-CNN for the instance segmentation of natural objects \parencite{moshkovTesttimeAugmentationDeep2020}, and U-Net for cell detection and morphometry (\cite{ronnebergerUNetConvolutionalNetworks2015}; \cite{falkUNetDeepLearning2019}).
Many variants and combinations have also been investigated, such as combining two CNNs with the watershed algorithm \parencite{luxCellSegmentationCombining2020}, and using a similar architecture to U-Net, called Hourglass network \parencite{payerSegmentingTrackingCell2019}. However, less attention is attracted on the association stage, and many algorithms simply adopt bipartite matching\parencite{ciaparroneDeepLearningVideo2020}.

\subsection{Our Approach}

To explore more possibilities of matching methods, we will adopt the second manner and concentrate on the association stage by assuming the detection stage has been done and the segmentation accuracy is accurate enough. The assumption is reasonable when the cells to be segmented are pretty regular and can be segmented with modern computer vision techniques such as the watershed algorithm for overlapped circular shapes \parencite{szeliskiComputerVision2011}.

The widely-used bipartite matching is not restricted to two consecutive frames as in \textcite{padfieldCoupledMinimumcostFlow2011}, but also can be used among tracklets, such as the second LAP in \textcite{jaqamanRobustSingleparticleTracking2008}.
However, the distance defined in the bipartite matching involves only two frames (or tracklets). It would fail in some cross-path situations, as shown in Proposition \ref{prop:cross}, we try to improve the accuracy of the bipartite matching with an acceptable additional computational cost.

The global linking also suffers the same shortcoming as the bipartite matching. The cost function defined for each edge in Figure \ref{fig:mincost2} cannot involve information from the previous frame since the previous matching also needs to be determined. For example, suppose we want to define the cost from $O_4$ to $O_8$, it will be more informative if we know which one linked to $O_4$ since a smaller distance between the first two frames generally tends to imply a small distance between the successive two frames, but the matching between the first two frames are also to be optimized.

Due to these limitations, we propose tripartite matching, which defines a target function using three frames and optimize it globally. Specifically, the target function is the velocity difference instead of the distance difference in bipartite matching and global linking. It is important to note that the treatment of velocity is quite different from other work. Several existing methods have incorporated the velocity, but they either ignore the velocity across the matching pairs or heavily depend on the quality of initialization. In the language of the solution space that we will discuss, they often miss some high-quality parts of the solution space and hence would easily lead to suboptimal solutions.

As a representative method considering velocity, \textcite{xingMultiobjectTrackingOcclusions2009} firstly generate tracklets by some filter methods and then perform bipartite matching on the tracklets. The lengths of tracklets are long enough to calculate the velocity. Then the velocity within each tracklet, together with the position and other potential shape features, are put together to calculate the Euclidean distance between two tracklets. However, the velocity from the first tracklet, the left side of bipartite matching, to the second tracklet, the right side of bipartite matching, cannot be included. In other words, only the within-tracklet velocity has been counted, and the between-tracklet velocity is ignored. Our approach would consider the velocity across each frame in the matching.

As another representative method considering velocity, \textcite{milanContinuousEnergyMinimization2014} also adopt two steps, where the first step aims to initialize a complete trajectory by some simple tracking methods such that the velocity can be calculated, then optimize their defined energy function, which consists of the velocity difference. They indeed consider the velocity for each frame, but the calculation would depend on the initialization. In our framework, we directly optimize the target function that involves the velocity. For computational efficiency, we will construct a reduced space with the help of another simple tracking method, similar to the first step in \textcite{milanContinuousEnergyMinimization2014}, but such a method merely aims to reduce the search space instead of providing a determined and complete trajectory.

As a special case, \textcite{vallottonTritrackFreeSoftware2013} do not explicitly consider the velocity, but they propose a three-frame tracker which implicitly uses the velocity information. However, they do not optimize in a global way since the tracker solves the matching between every three frames, and it constructs the solution by enumerating all possible combinations of objects in three frames, which seems computationally extensive. Although they introduce an additional constant threshold to filter out objects with a distance larger than this given constant, the threshold would cost more effort and is not adaptive for different objects and different frames.

In general, the cells in cell tracking are more crowded than in other applications. For example, \textcite{milanContinuousEnergyMinimization2014} validate their multiple human tracking algorithms on the visual surveillance videos PETS2010 \parencite{maskaBenchmarkComparisonCell2014}, the number of pedestrians in each frame can be less than 10. Although there is a scenario where 42 pedestrians are walking simultaneously, nearly all people walk on the same road and even in the same direction. However, there are always around 50 cells per frame in our real cell video, and their directions can be arbitrary. The computational burden of tripartite matching increases along the number of objects to be tracked, and we try to reduce the computational cost without sacrificing much accuracy.

Section \ref{sec:model} proposes a tripartite model in a probabilistic framework.
Section \ref{sec:method} introduces the dynamic programming with reduced search space for solving the tripartite model. Section \ref{sec:simulations} presents extensive simulations to compare our proposed approach with several popular methods, and Section \ref{sec:ctc} evaluates the approaches on some public datasets in the Cell Tracking Challenge.
In Section \ref{sec:real-cell-video}, we apply the tracking algorithms on a real cell video, and show the superiority of our proposed method over the distance method with some case studies.
Finally, we discuss some limitations and extensions of our work in Section \ref{sec:conclusion}. The detailed mathematical formulations, necessary proofs, and technical implementations are given in the Appendix.

\section{Model}\label{sec:model}

\subsection{Velocity Model (Tripartite Model)}

Let $Z_k=(Z_{k1}, Z_{k2},\ldots, Z_{kn_k})$ be a state vector of $n_k$ objects at frame $k$, each of which consists of the position
and velocity. We label each of the $n_k$ objects with a unique integer from 1 to $n_k$. The labelling of objects within a framework can be arbitrary as long as the labelling is fixed afterwards. For object $i$ in the frame $k$, let $(x_{ki},y_{ki})$ be the position, and $(\dot x_{ki},\dot y_{ki})$ be the velocity. Then the state vector becomes $Z_{ki}=(x_{ki},y_{ki},\dot x_{ki},\dot y_{ki})$.

\begin{remark}
  If the state vector consists of only the position, i.e., $Z_{ki} = (x_{ki}, y_{ki})$ for object $i$ at the frame $k$, the resulting model, called position model or bipartite model, would be equivalent to the distance-based bipartite model under certain conditions, see Supplementary Material for more details. Furthermore, the state vector can be extended by including various descriptors for the appearance of the objects, such as the color histogram, the histogram of oriented gradient (HOG), and the region covariance matrix \parencite{luoMultipleObjectTracking2014}.
\end{remark}

Consider two matched objects $Z_{ki}$ and $Z_{k+1,j}$ in two consecutive frames $k, k+1$. 
By simple relationship between velocity and displacement, we have
\begin{equation}
  \begin{bmatrix}
    x_{k+1,j}\\
    y_{k+1,j}\\
    \dot x_{k+1,j}\\
    \dot y_{k+1,j}\\
  \end{bmatrix} =
  \begin{bmatrix}
    1 & 0 & \Delta t & 0\\
    0 & 1 & 0 & \Delta t\\
    0 & 0 & 1 & 0\\
    0 & 0 & 0 & 1
  \end{bmatrix}
  \begin{bmatrix}
    x_{ki}\\
    y_{ki}\\
    \dot x_{ki}\\
    \dot y_{ki}
  \end{bmatrix}
   +
  \begin{bmatrix}
    \Delta t & 0\\
    0 & \Delta t\\
    1 & 0\\
    0 & 1
  \end{bmatrix}
  \begin{bmatrix}
    \epsilon_{i,\dot x}\\
    \epsilon_{i,\dot y}
  \end{bmatrix}
  \equiv FZ_{ki} + G\epsilon_i
  \label{eq:motion}
\end{equation}
where $\epsilon_{i, \dot x}, \epsilon_{i,\dot y}$ are the residual velocity along the $x$-axis and $y$-axis respectively, and multiplying $\Delta t$ yields the residual position.
We make the velocity change smoothly by assuming $\epsilon_i\sim N(0,\Sigma_k)$, then
\begin{equation}
Z_{k+1,j}\mid Z_{ki}\sim N(FZ_{ki}, G\Sigma_k G')\,.\label{eq:matrix_state_vector}
\end{equation}
Note that $\rank(G\Sigma_k G')\le \rank(G) = 2$, which implies that $N(0,G\Sigma_k G')$ is not absolutely continuous and has no probability density function, and hence we cannot put the position and velocity in a state simultaneously.
To avoid this problem, we multiply \eqref{eq:matrix_state_vector} by
\begin{equation}
  H_v =
  \begin{bmatrix}
    0 & 0 & 1 & 0\\
    0 & 0 & 0 & 1
  \end{bmatrix}\,,
\end{equation}
then
\begin{equation}
  H_vZ_{k+1,j}\mid Z_{ki}\sim N(H_vFZ_{ki}, H_vG\Sigma_k G'H_v')\label{eq:vel_model}\,.
\end{equation}
Specifically, we have
\begin{equation}
  \begin{bmatrix}
    \dot x_{k+1,j}\\
    \dot y_{k+1,j}
  \end{bmatrix}
  =
  \begin{bmatrix}
    \dot x_{ki}\\
    \dot y_{ki}
  \end{bmatrix}
  +
  N(0,\Sigma_k)\,.\label{eq:sub_vel}
\end{equation}
\begin{remark}
  There is another equivalent way to obtain \eqref{eq:sub_vel}.
  Multiplying \eqref{eq:matrix_state_vector} by
  \begin{equation}
    H_p =
    \begin{bmatrix}
      1 & 0 & 0 & 0\\
      0 & 1 & 0 & 0
    \end{bmatrix}
  \end{equation}
  yields
  \begin{equation}
    H_pZ_{k+1,j}\mid Z_{ki}\sim N(H_pFZ_{ki}, H_pG\Sigma_k G'H_p')\label{eq:pos_model}\,,
  \end{equation}
  that is,
  \begin{equation}
    \begin{bmatrix}
      x_{k+1,j}\\
      y_{k+1,j}
    \end{bmatrix}
    =
    \begin{bmatrix}
      x_{ki} + \dot x_{ki}\Delta t\\
      y_{ki} + \dot y_{ki}\Delta t
    \end{bmatrix}
    +
    N\left(0,\Delta t^2\Sigma_k\right)\label{eq:sub_pos}\,.
  \end{equation}
  The equivalence between \eqref{eq:sub_pos} and \eqref{eq:sub_vel} comes from the fact that
  $\dot x_{k+1,j}\Delta t=x_{k+1,j}-x_{ki}$ and $\dot y_{k+1,j}\Delta t=y_{k+1,j}-y_{ki}$.
\end{remark}

It follows that $Z_k$ forms a Markov chain, i.e.,
$$
\Pr(Z_{k+1}\mid Z_{k}, \ldots, Z_1) = \Pr(Z_{k+1}\mid Z_k)\,.
$$

\begin{remark}
  Unlike the probabilistic framework used in the stochastic filter methods, which treats the observations as random and imposes (Gaussian) noises, the object positions are already given without uncertainty since we have assumed the cell segmentation stage has been done.
  Another difference is that their probabilistic framework usually results in online tracking methods since they iteratively perform the updating step and the predicting step, while the probabilistic method developed here is for offline use.
\end{remark}

\subsection{Matching Vector}
Each trajectory can be described as association/matching between every two adjacent frames, although the estimation of matching does not necessarily involve only these two frames. Formally, to match frame $k$ with frame $k+1$, where the number of objects are $n_k$ and $n_{k+1}$. Let $M_{k,k+1}$ be a $n_k$-vector,
\begin{equation}
  M_{k,k+1}[i] =
  \begin{cases}
      j &\text{ if }\text{object $i$ in frame $k$ corresponds to object $j$ in frame $k+1$}\\
      -1 &\text{ if }\text{object $i$ leaves out of the visible region}
  \end{cases}\,,
\end{equation}
where $i \in \{1,2,\ldots,n_k\}$ is the index at frame $k$, and $j\in \{1,2,\ldots,n_{k+1}\}$ is the index at frame $k+1$, and indexes across different frames are independent.

\begin{remark}
Here is an alternative way to interpret the matching vector $M_{k,k+1}$. Let $\M$ be a $n_k\times n_{k+1}$ binary matrix, whose entries are either 0 or 1. If $\M[i, j] = 1$, then object $i$ in frame $k$ is matched to object $j$ in frame $k+1$. For the object in the $k$-th frame,
each object either disappears or stays in the visible region, i.e.,
$$
\sum_{j=1}^{n_{k+1}}\M[i, j] \le 1\quad \forall i=1,\ldots,n_k\,,
$$
and similarly, for the object in the $k+1$ frame, the object either just appears or has existed in the previous frame,
$$
\sum_{i=1}^{n_k}\M[i, j] \le 1\quad \forall j=1,\ldots,n_{k+1}\,.
$$
There is a one-to-one correspondence between $M$ and $\M$. More specifically,
$$
M_{k,k+1}[i] = \begin{cases}
\argmax_j \M[i, j]  & \text{ if }\sum_{j=1}^{n_{k+1}}\M[i,j] = 1\\
-1 & \text{ if }\sum_{j=1}^{n_{k+1}}\M[i, j] = 0
\end{cases}
$$
and
$$
\M[i, j] = \begin{cases}
1 & \text{ if }M_{k,k+1}[i] = j\\
0 & \text{ otherwise }
\end{cases}\,.
$$
\end{remark}

In 2D situations, where no objects disappear or appear from the middle, the matching vectors $M=(M_{12},\ldots, M_{f-1,f})$ uniquely determine the trajectories of all cells.

\begin{remark}
  In some cell tracking tasks, there might be cell division and merging. Moreover, in 3D problems, objects can enter into or leave the visible region in the middle due to vision depth or occlusion.
\end{remark}

Let $[n_k]=\{1,2,\ldots,n_k\}$ be the index of objects at frame $k$, and suppose there are $d$ disappeared objects, then a typical matching vector can be
$$
M_{k,k+1}^{d,i}=\left[\underbrace{e_1,e_2,\ldots,e_{n_k-d}}_{\text{choose from $[n_{k+1}]$}}, \underbrace{-1,\ldots,-1}_{d}\right], e_1 < e_2 <\cdots < e_{n_k-d},i=1,\ldots,\binom{n_{k+1}}{n_k-d}\,,
$$
where $\max(0, n_k-n_{k+1})\le d\le n_k$ and $i$ indexes the choice of picking $n_k-d$ elements from $[n_{k+1}]$. Any permutation of $M_{k,k+1}^{d,i}$ would be another matching vector, denoted as $\bar \pi(M_{k,k+1}^{d,i})$. All possible permutations constitute the whole space of the matching vector,
\begin{equation}\label{eq:Dk}
  D_k=\cup_d\cup_i\bar\cP(M_{k,k+1}^{d,i}) = \cup_d\cup_i\bar\cP(\bar\pi(M_{k,k+1}^{d,i}))\,.
\end{equation}
and
$$
\bar\cP(M_{k,k+1}^{d,i})\cap \bar\cP(M_{k,k+1}^{d',j}) = \emptyset\qquad \forall i\neq j \text{ or }d\neq d'\,,
$$
where $\bar\cP(M_{k,k+1}^{d,i})$ consists of all possible permutations (including the identity permutation) of $M_{k,k+1}^{d,i}$, and $\bar\pi$ is one particular permutation.
Moreover, $\bar\pi$ can be further decomposed as
$$
\bar\pi(M_{k,k+1}^{d,i}) = \pi(\tau_j(M_{k,k+1}^{d,i}))\,,
$$
where $\tau_j, j = 1,2,\ldots, \binom{n_k}{d}$ determines the positions of disappeared the element $-1$ and $\pi$ permutes the remaining non-disappeared elements. Thus,
$$
D_k=\cup_d\cup_i\bar\cP(\bar\pi(M_{k,k+1}^{d,i})) = \cup_d\cup_i\cup_j\cP(\tau_j(M_{k,k+1}^{d,i}))\,,
$$
where $\cP(M_{k,k+1})$ is the set constituted by all \emph{partial} permutations (including the identity permutation) of matching vector $M_{k,k+1}$, where \emph{partial} means to permute only the non-disappeared elements.

\subsection{Likelihood Function}

We formulate the optimal matching vectors $M=(M_{12},\ldots,M_{f-1,f})$ as the point in the space $D_1\times \cdots\times D_{f-1}$ which maximizes the likelihood of the state vector $Z = (Z_1,\ldots,Z_f)$,
\begin{align}
M^\star & = \argmax_{M\in D_1\times \cdots\times D_{f-1}} P(Z_1,\ldots, Z_f\mid M)\\
&=\argmax_{M\in D_1\times \cdots\times D_{f-1}} P(Z_1)\prod_{k=2}^f P(Z_k\mid Z_{k-1},M)\\
&=\argmax_{M\in D_1\times \cdots\times D_{f-1}} \prod_{k=2}^f P(Z_k\mid Z_{k-1},M)\label{eq:Mstar_x}\,,
\end{align}
where $f$ is the total number of frames.

\begin{proposition}\label{prop:global_vs_local}
  For the state vector $Z_k$ satisfying
  \begin{equation}
  P(Z_k\mid Z_{k-1}, M) = P(Z_k\mid Z_{k-1}, M_{k-1, k})\label{eq:cond_pairs}\,,
  \end{equation}
  solving the global matching $M$ is equivalent to solving the pairwise matching $M_{k-1,k}$ separately.
\end{proposition}

Note that the velocity is computed as the displacement from the previous frame to the current frame, divided by the time interval. It follows that
\begin{equation}
  P(Z_{k+1}\mid Z_{k},M) =
  P(Z_{k+1}\mid Z_{k},M_{k,k+1}, M_{k-1,k}) =
  P(\bfx_{k+1}\mid \bfx_{k},\bfx_{k-1},M_{k,k+1}, M_{k-1, k})\,\label{eq:tri_ll}
\end{equation}
involves three frames and does not satisfy Proposition \ref{prop:global_vs_local}, we also call it a tripartite model whose computation complexity is much larger than the bipartite model.
Since we allow objects to disappear/appear at an arbitrary frame, the formulation of \eqref{eq:tri_ll} would depend on the existence statuses of objects on three consecutive frames. For example, it might exist at the first two frames and then disappear, or appear at the second frame and stay in the visible region. The detailed calculations for each object existence status refer to Supplementary Material.
\begin{remark}
  The position model (bipartite model) satisfies Proposition \ref{prop:global_vs_local}, and hence it can be efficiently solved like the distance-based bipartite matching of \textcite{padfieldCoupledMinimumcostFlow2011}.
\end{remark}

\section{Method}\label{sec:method}

Rewrite the likelihood function \eqref{eq:Mstar_x} for the tripartite model as
\begin{align}
\sum_{k=2}^f \log P(Z_{k}\mid Z_{k-1}, M)&=\log P(\bfx_2\mid \bfx_1,M_{12}) + \sum_{k=3}^f\log P(\bfx_{k}\mid \bfx_{k-1}, \bfx_{k-2}, M_{k-1,k}, M_{k-2,k-1})\\
&\triangleq h_1(M_{12}) +\sum_{k=2}^{f-1}h_k(M_{k-1,k}, M_{k,k+1})\label{eq:tri_obj_all}\,,
\end{align}

This chain structure, where each matching vector $M_{k-1, k}$ is in two neighboring functions $h_{k-1}$ and $h_k$, implies that the optimal solution of the optimization problem for video with first $k$ frames depends on the optimal solution of the problem for video with first $k-1$ frames. Thus we can break the original optimization problem into simpler sub-problems in a recursive manner, which indicates that the dynamic programming is a natural choice \parencite{cormenIntroductionAlgorithms2009}.

\subsection{Dynamic Programming}

The dynamic programming can be used to maximize \eqref{eq:tri_obj_all} as follows:
\begin{enumerate}
  \item Define
  $$
  m_1(x) = h_1(x)\quad \forall x\in D_1
  $$
  and
  $$
  m_2(x) = \max_{M_{12}\in D_1} m_1(M_{12})+h_2(M_{12}, x)\quad \forall x \in D_2\,.
  $$
  \item Recursively compute the function
  \begin{equation}
  m_k(x) = \max_{M_{k-1,k}\in D_{k-1}} \{m_{k-1}(M_{k-1,k}) + h_k(M_{k-1,k}, x)\}\quad \forall x\in D_k\label{eq:dp2}
  \end{equation}
  for $k=3,4,\ldots,f-1$.
  \item The optimal value is attained by
  $$
  \max_{M_{f-1,f}\in D_{f-1}}m_{f-1}(M_{f-1,f})\,.
  $$
\end{enumerate}

Then trace backward to find out which $M$ gives rise to the global maximum.

\begin{enumerate}
  \item Let $\hat M_{f-1,f}$ be the maximizer of $m_{f-1}(x)$, i.e.,
  $$
  \hat M_{f-1,f} = \argmax_{M_{f-1,f}\in D_{f-1}}m_{f-1}(M_{f-1,f})\,.
  $$
  \item For $k=f-2,\ldots,2$, let
  $$
  \hat M_{k,k+1} = \argmax_{M_{k,k+1}\in D_k}\{m_k(M_{k,k+1}) + h_{k+1}(M_{k,k+1}, \hat M_{k+1,k+2})\}
  $$
  \item For the first term,
  $$
  \hat M_{12} = \argmax_{M_{12}}\{h_1(M_{12})+h_2(M_{12},\hat M_{23})\}\,.
  $$
\end{enumerate}

\subsection{Reduction of search space}

Recall the definition \eqref{eq:Dk} of the search space $D_k$ for $M_{k,k+1}$, whose size can be calculated as
$$
\vert D_k\vert
= \sum_d\binom{n_k}{d}\binom{n_{k+1}}{n_k-d}(n_k-d)!
= \sum_d \binom{n_k}{d}\frac{n_{k+1}!}{(n_{k+1}-n_k+d)!}\,,
$$
in which firstly we determine the location of disappeared cells (i.e., element $-1$) from all $\binom{n_k}{d}$ possibilities and the candidates of non-disappeared cells from all $\binom{n_{k+1}}{n_k-d}$ possible choices, and then perform a permutation on the non-disappeared cells.
It follows that the whole complexity of the dynamic programming would be
$$
O\left(\sum_{k=1}^{f-1}\vert D_k\vert^2\right)\,.
$$
\begin{proposition}\label{prop:raw_complexity}
  The size of the space $D_k$ is $\Omega(\min(n_k,n_{k+1})!)$, and hence the complexity of \eqref{eq:dp2} is $\Omega(\min(n_{k-1}, n_{k})!\cdot\min(n_{k}, n_{k+1})!)$. Further assuming $n_k \sim N$, then the complexity is simplified to $\Omega((N!)^2)$.
\end{proposition}

To reduce the computational complexity without sacrificing much performance, consider the search space consisting of the variants of a bipartite matching vector, i.e., some proper permutations of the matching vectors. The bipartite matching vector can be chosen as the one obtained by \textcite{padfieldCoupledMinimumcostFlow2011}'s bipartite matching, or any other matching vectors derived from the bipartite model. The intuition is that we can correct the mismatches caused by the crossed paths in the bipartite model, as discussed in Proposition \ref{prop:cross}.

\begin{figure}[H]
  \centering
  \begin{subfigure}{0.33\textwidth}
    \includegraphics[page=1,width=\textwidth]{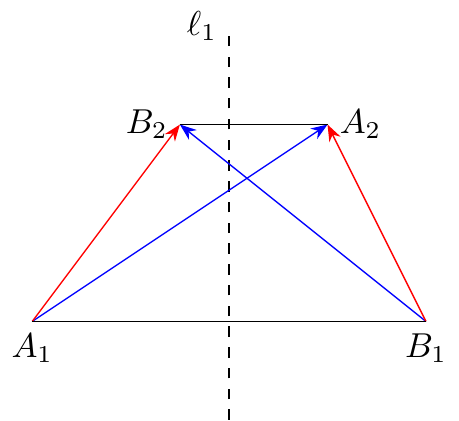}
    \caption{}
    \label{fig:cross_case1}
  \end{subfigure}%
  \begin{subfigure}{0.33\textwidth}
    \includegraphics[page=2,width=\textwidth]{figs/cross.pdf}
    \caption{}
    \label{fig:cross_case2}
  \end{subfigure}%
  \begin{subfigure}{0.33\textwidth}
    \includegraphics[page=3,width=\textwidth]{figs/cross.pdf}
    \caption{}
    \label{fig:cross_case3}
  \end{subfigure}%
  \caption{Diagram of crossed paths in different fashions.}
  \label{fig:cross}
\end{figure}

\begin{proposition}\label{prop:cross}
  Suppose two paths $A_1A_2$ and $B_1B_2$ cross, where the subscripts denote the time frame. Let $\ell_1, \ell_2$ be the vertical bisector of $A_1B_1$ and $A_2B_2$, respectively. For the bipartite model equipped with any cost function $\varphi$ whose value depends only on the distance such that $\varphi(\bfx, \bfx') = \varphi(\Vert \bfx - \bfx'\Vert)$, where $\bfx, \bfx'$ denotes the coordinates of two cells,
  \begin{itemize}
    \item it would mismatch if $\ell_1$ separates $A_2$ and $B_2$, i.e., one on the left of $\ell$, and another on the right, as shown in Figure \ref{fig:cross_case1};
    \item it would mismatch if $\ell_1$ cannot separate $A_2$ and $B_2$, but $\ell_2$ can separate $A_1$ and $B_1$, as shown in Figure \ref{fig:cross_case2};
    \item the matching depends on the cost function if neither $\ell_1$ nor $\ell_2$ separate two cells, as shown in Figure \ref{fig:cross_case3}.
  \end{itemize}
  Furthermore, if the cost function is taken as the square of distance, $\varphi(\bfx,\bfx') = \Vert \bfx-\bfx'\Vert_2^2$, then the model would always fail.
\end{proposition}

Take a toy example for illustration, suppose the bipartite model returns the mismatches $(A_1\rightarrow B_2, B_1\rightarrow A_2)$ for cell $A$ and $B$, where the subscripts denote the frame index, i.e., the predicted matching vector is $\hat M_{12}=[2,1]$, while the truth is $M_{12}=[1,2]$. If we permute any two elements in $\hat M_{12}$, where the possible permutations include the identity permutation, i.e., $\hat M_{12}$ itself, then we construct a search space $\{[2,1], [1,2]\}$, which contains the truth and it might be identified by the tripartite model \eqref{eq:sub_vel}. Note that the whole search space is $\{[2, 1], [1, 2], [1], [2], []\}$, which is much larger than the reduced space, but we still can get correct matching results without enumerating all possible cases from the whole search space.

Formally, the bipartite matching vector can give us some hints to construct the reduced space. First of all, it provides us an estimate of the number of disappeared cells, $d^\star$, so the range of $d$ can be restricted to
$$
N(d^\star) = [\max(0, n_k-n_{k+1}), n_k]\cap [d^\star-\delta, d^\star + \delta]\,,
$$
where $\delta$ is a tuning parameter that controls the size of reduced search space. Secondly, exchanging some elements of the matching vector can recover the truth, as illustrated in the above toy example. Given the number of disappeared cells $d$, one can get a fixed-$d$ bipartite matching $\hat M_{k,k+1}^d$ by the algorithm discussed in Supplementary Material, which implies that
$$
\hat M_{k,k+1}^d = \pi^\star(\tau_{j^\star}(M_{k,k+1}^{d,i^\star}))\,,
$$
where $i^\star$ indicates the particular choice of sub-vector from $[n_{k+1}]$, and $j^\star$ determines the locations of disappeared cells, i.e., the components with value $-1$, while $\pi^\star$ represents the particular permutation on the non-disappeared cells. Moreover, we exchange only one pair, $\cP_1$, including the identity permutation. Now it is ready to define the reduced space as
$$
\tilde D_k=\bigcup_{d\in N(d^\star)}\bigcup_{i=i^\star}\bigcup_{j=j^\star}\cP_1(\pi^\star(\tau_j(M_{k,k+1}^{d,i})))=\bigcup_{d\in N(d^\star)}\cP_1(\hat M_{k,k+1}^d)\,.
$$

\begin{proposition}\label{prop:reduced_complexity}
  The size of the reduced space $\tilde D_k$ is $O((\delta+1/2)n_k^2)$, and hence the complexity of \eqref{eq:dp2} is $O((\delta+1/2)^2n_{k-1}^2n_k^2)$. Further assuming $n_k\sim N$, the the complexity becomes $O((\delta+1/2)^2N^4)$.
\end{proposition}

As for the computation complexity of solving the bipartite model by the min-cost flow algorithm, several diverse implementations have different computation complexities. A common one as analyzed in \textcite{padfieldCoupledMinimumcostFlow2011} is $O(n_k^3\log n_k)$. It is no surprise that the complexity of our proposed DP with reduced space would be higher, but it seems comparable for moderate $N$. However, the memory allocations in dynamic programming cannot be negligible like in min-cost flow algorithm because we need to store all maximum functions $m_k(x)$, and the cost (or score) evaluations $h_k$ are much expensive than the bipartite models that based only on the distance. Fortunately, we can optimize the memory allocations and cost evaluations by Proposition \ref{prop:evalcost}.
Specifically, if we have calculated $h_{k-1}(M_{k-1,k}, \hat M^d_{k,k+1})$, then we can quickly obtain $h_{k-1}(M_{k-1,k}, M_{k,k+1})$ for all $M_{k,k+1}\in \cP_1(\hat M^d_{k,k+1})$.
Nevertheless, the computational cost would be higher than the bipartite model, but we will observe that it rewards much better performance in Section \ref{sec:simulations}.

\begin{proposition}\label{prop:evalcost}
  The difference between $h_{k-1}(M_{k-1,k}, \hat M^d_{k,k+1})$ and $h_{k-1}(M_{k-1,k}, M_{k,k+1}), M_{k,k+1}\in \cP_1(\hat M^d_{k,k+1})$ involves only the cost of the exchanged pair.
\end{proposition}

\subsection{Estimation of $\Sigma_k$}\label{sec:sigma}

\begin{figure}[H]
\sbox\twosubbox{%
  \resizebox{\dimexpr\textwidth-1em}{!}{%
    \includegraphics[page=2,height=2cm]{figs/sigma.pdf}%
    \includegraphics[page=1,height=2cm]{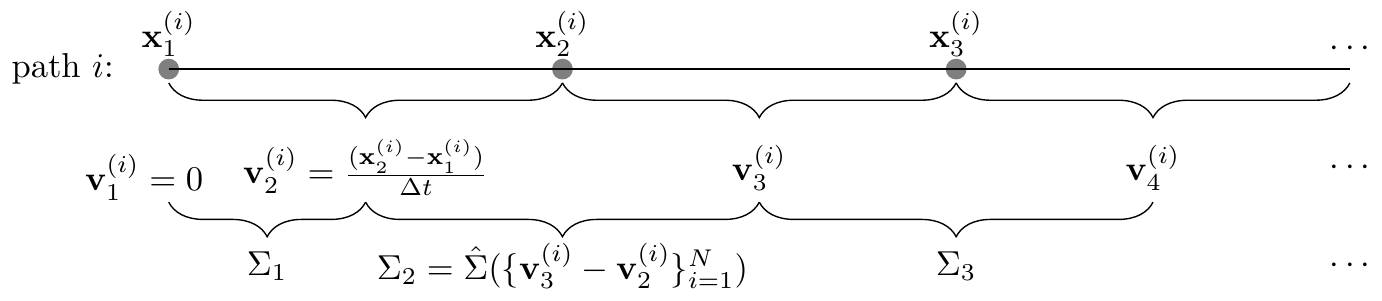}%
  }%
}
\setlength{\twosubht}{\ht\twosubbox}
\centering
\subcaptionbox{Paths.\label{fig:paths}}{%
  \includegraphics[page=2,height=\twosubht]{figs/sigma.pdf}%
}\,
\subcaptionbox{Estimate $\Sigma_k$.\label{fig:estsigma}}{%
  \includegraphics[page=1,height=\twosubht]{figs/sigma.pdf}%
}
\caption{Schematic diagram for estimating $\Sigma_k$. (\emph{a}) The points on the same row are the same object on four consecutive frames and are denoted by the same color. The lines linking the points denote the estimated paths by the association algorithm. (\emph{b}) The estimation procedure for the covariance is based on these estimated paths.}
\end{figure}

In practice, the true $\Sigma_k$ is unknown, and we need to estimate it. Since we have performed the bipartite model first, then the sample covariance of the velocity difference calculated from the estimated paths would be a natural estimator, as illustrated in Figure \ref{fig:estsigma}, where $\hat\Sigma$ denotes the sample covariance based on all $N$ paths and path $i$ might be one of the paths shown in Figure \ref{fig:paths}.

Without particular reason showing the movement along $x$-axis and $y$-axis are correlated and heterogeneous, we will prefer to take $\Sigma_k=\sigma_k^2I$, and then the velocity difference along $x$-axis and $y$-axis can be pooled to get an overall estimate of $\sigma_k$. Furthermore, if the velocity variations among all frames are assumed to be the same, we can obtain the pooling estimate $\hat\sigma$, not restricted to some particular frame $k$. The following Proposition \ref{prop:sigma} tells us the pooled estimate $\hat\sigma$ over all frames can be chosen arbitrarily if we take the tuning parameter $\delta = 0$. On the other hand, if we take different $\sigma_k$ for different frame pairs, $\sigma_k$ (or more accurately, $\frac{1}{\sigma_k^2}$) can be interpreted as the weights of the velocity differences in $\log h_k(M_{k,k+1}\mid M_{k-1,k},\sigma_k)$, refer to more details in Supplementary Material. It imposes a smaller weight for larger $\sigma_k$, which is reasonable and helpful for matching since higher $\sigma_k$ tends to imply more uncertainty.

\begin{proposition}\label{prop:sigma}
  For any $\Sigma_k^{(1)}=\sigma^{(1)}_kI, \Sigma_k^{(2)}=\sigma^{(2)}_kI, \forall k=1,\ldots,f-1$,
  $$
  \argmax_{M_{12}\in \cP_1(\hat M_{12}^d)} h_1(M_{12}\mid \sigma_1^{(1)}) = \argmax_{M_{12}\in \cP_1(\hat M_{12}^d)}h_1(M_{12}\mid\sigma_1^{(2)})\,,
  $$
  and
  $$
  \argmax_{M_{k,k+1}\in \cP_1(\hat M_{k,k+1}^d)} h_k( M_{k,k+1}\mid M_{k-1,k},\sigma_k^{(1)}) = \argmax_{M_{k,k+1}\in \cP_1(\hat M_{k,k+1}^d)} h_k(M_{k,k+1}\mid M_{k-1,k}, \sigma_k^{(2)})\,.
  $$
  But it is NOT necessary to hold
  $$
  \argmax_{M_{k,k+1}\in \tilde D_k} h_k(M_{k,k+1}\mid M_{k-1,k}, \sigma_k^{(1)}) = \argmax_{M_{k,k+1}\in \tilde D_k} h_k(M_{k,k+1}\mid M_{k-1,k}, \sigma_k^{(2)})\,.
  $$
  Furthermore, suppose that $\sigma^{(1)}_k=\sigma^{(1)}, \sigma^{(2)}_k=\sigma^{(2)},\forall k$, the whole path matching would be exactly the same.
\end{proposition}

Since each object's movement is assumed to be independent, and their velocity difference follows the same distribution $N(0, \Sigma_k)$, the sample covariance of the velocity difference would be a consistent estimator given the unknown authentic trajectories. However, in practice, the estimated sample covariance is based on the predicted trajectories obtained by the bipartite model, which would have some mismatches, such as the arrow lines connecting different colored points in Figure \ref{fig:paths}, where the (hidden) horizontal lines linking the same colored points are the unknown authentic paths. Although the bipartite model can conduct the matching pairwisely as shown in Proposition \ref{prop:global_vs_local}, i.e., the mismatches between every two frames are independent, the path error is accumulated. For example, all paths in the first two frames in Figure \ref{fig:paths} are correct, but only one path is correct in the first three frames, and finally, none path is correct in all four frames. Consequently, the error of the estimation $\hat\Sigma_k$ would be accumulated, and it would become more and more overestimated along with the time frame since the mismatches usually cause the velocity difference to more disperse.

On the other hand, our main interest is the matching performance instead of the consistency estimator of $\Sigma_k$. It would be acceptable if the effect of $\hat\Sigma_k$ is minimal or even negligible, as discussed in Proposition \ref{prop:sigma}. The following simulations will investigate the actual impact of $\hat\Sigma_k$ on the matching performance.

\section{Simulations}\label{sec:simulations}

It will take much effort to compare different methods' performance on the real cell video since we do not have any labeled trajectories, and the segmentation qualities might harm the tracking accuracy. This section provides a platform for comparing the tracking accuracy in isolation from the segmentation performance. Extensive simulations have been conducted on our proposed method and another four popular association methods mentioned in Section \ref{sec:intro}, \textcite{padfieldCoupledMinimumcostFlow2011}'s Bipartite matching solved by the Minimum-Cost Flow, abbreviated as BMCF, \textcite{zhangGlobalDataAssociation2008}'s Global association solved by the Minimum-Cost Flow too, abbreviated as GMCF, \textcite{jaqamanRobustSingleparticleTracking2008}'s LAP, and \textcite{magnussonGlobalLinkingCell2015}'s graph-based global linking, also known as Baxter algorithm.

\subsection{Setting}

\begin{figure}[H]
  \centering
  \includegraphics[width=\textwidth]{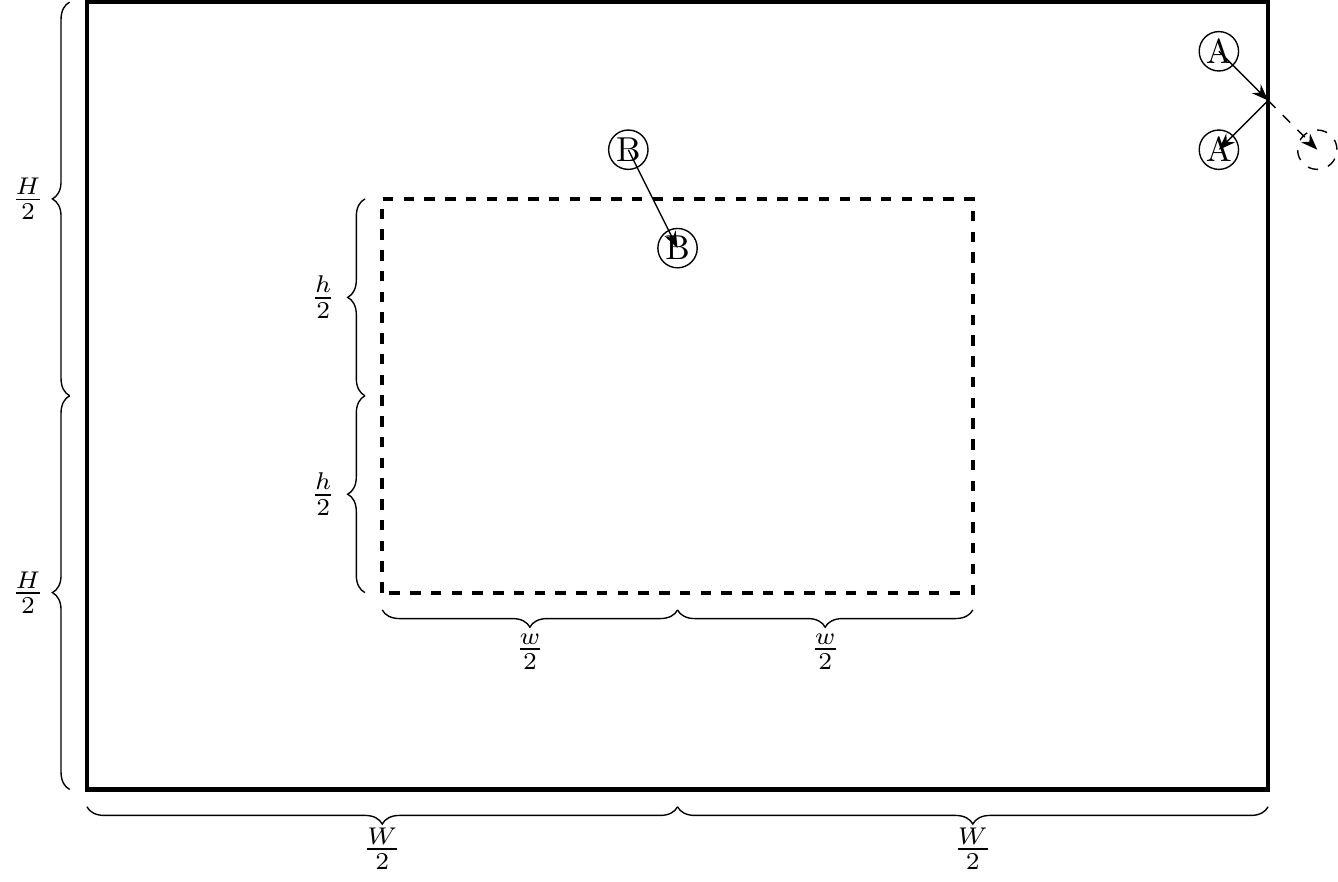}
  \caption{Schematic diagram of the simulated cell video. The solid rectangle with size $W\times H$ represents the closed region, while the $w\times h$ dashed rectangle is visible. The circle $A$ is a cell that hits the boundary and would be reflected back, while cell $B$ can freely move into the visible region.}
  \label{fig:wh}
\end{figure}

Suppose we have a closed region, such as the solid rectangle in Figure \ref{fig:wh}, where cells can move inside freely except that they cannot move out or into this closed region. If a cell hits the boundary, such as cell A in Figure \ref{fig:wh}, it will be reflected along the solid arrowed line.
To simulate a cell video, we need to choose a sub-region as the camera's visible region first, and then we can begin photographing by focusing on such a visible region. For simplicity, suppose the visible region is located precisely in the center, i.e., the dashed rectangle inside the big solid one. Let $W, H$ be the width and height for the solid rectangle, respectively, while $w, h$ are the width and height for the dashed one, then the scaling factors are defined as $W/w$ and $H/h$. In contrast to the reflection on the closed region boundary, cells can enter into or leave the visible region from the dashed border. Hence, the simulations allow cells to disappear or appear.

Note that the total number of cells in the closed region is constant if there is no cell merging and splitting, while the number of cells in the visible region would always be changing since cells can enter into (appear) or leave out (disappear) from the visible region. It can be expected that the ratio between the numbers of cells in the visible region and the whole closed region is rough $\frac{wh}{WH}$ if the cells distribute uniformly. As an initialization, generate $\frac{WH}{wh}N_0$ cells uniformly in the closed rectangle such that the expected number of cells in the visible region is $N_0$. In the following experiments, we fix $W/w = H/h = 5$ and take $w = 680, h=512$ to keep the same dimension as the ones of the image from the real cell video, and also fix the number of frames $f=50$.

Suppose the movements of each cell are independent, then we only need to focus on a single object's motion. It is natural to fix the time interval between two frames, say $\Delta t$. Then the generative model is
\begin{equation}
\begin{bmatrix}
  x_{k+1}\\
  y_{k+1}
\end{bmatrix}
=
\begin{bmatrix}
  x_k\\
  y_k
\end{bmatrix}
+
\begin{bmatrix}
  v_{xk}+\varepsilon_x\\
  v_{yk}+\varepsilon_y
\end{bmatrix}
\Delta t
\end{equation}
where $\varepsilon=[\varepsilon_x,\varepsilon_y]'\sim N(0,\sigma^2I)$, and for $k=2,3,\ldots$
$$
v_{xk}=\frac{x_{k}-x_{k-1}}{\Delta t}\quad v_{yk}=\frac{y_{k}-y_{k-1}}{\Delta t}
$$
while $v_{x1} = v_{y1} =0$, i.e., suppose the cells in the first frame are still.

\subsection{Metrics}
To assess the performance of the matching results, we consider the following metrics:
\begin{itemize}
  \item Pair accuracy: compare the accuracy between two consecutive frames.
  \item Whole Path accuracy: recover the path based on the matching vectors, then calculate the accuracy by comparing the predicted paths with the actual paths.
  \item Cumulative Path accuracy: stack the whole path accuracy for the sub-video constituted by the first $k$ frames, where $k=2,\ldots,f$.
\end{itemize}

The above accuracy can be precision (the proportion of correct predicted paths among the total amount of predicted paths) or recall (the percentage of correct predicted paths over the total amount of actual paths), or even the (weighted) harmonic mean of precision and recall,
$$
F_\beta = \frac{1+\beta^2}{\text{precision}^{-1} + \beta^2\text{recall}^{-1}}\,,
$$
which is called $F_\beta$ score \parencite{goutteProbabilisticInterpretationPrecision2005}. Since we want to measure comprehensively but do not have a preference on the recall and precision, just take $\beta = 1$ to use the balanced $F_1$ score.

If a predicted matching vector between two frames is precisely the same as the real matching vector, both the precision and the recall are 100\%, then we say the pair identity is 1; otherwise, the pair identity equals 0. Similarly, we can define the path identity, which takes 1 only when all paths are the same as the underlying truth. These binary quantities can also measure the matching performance, although more strict than the accuracy. Specifically, the accuracy measured by the (average) binary identity cannot distinguish the wrong paths with different amounts of mistakes. In general, we might not expect the path identity to be 1, but it is more likely for the pair identity to be 1, which is related to the following truth coverage.

The truth coverage aims to measure the efficiency of the reduced search space covering the true matching vector. Let $C(t)$ be the coverage status for matching frame $t$ and $t+1$. If $C(t)=1$, the true matching vector is included in the search space; otherwise, the search space excludes the truth. Now suppose we have independently conducted $N$ experiments under the same setting, define the average coverage rate as
$$
\bar C(t) = \frac 1N\sum_{i=1}^NC_i(t)\,,
$$
where $C_i(t)$ is the coverage status for experiment $i$ at frame $t$. The bipartite matching method, whose search space can be interpreted as the one-element space, consists of only its matching vector, then truth coverage status is exactly the pair identity. Since the reduced method's search space always contains the matching vector obtained by the corresponding bipartite method (here we adopt \textcite{padfieldCoupledMinimumcostFlow2011}'s BMCF), then the coverage rate must not be worse than the bipartite method. On the other hand, the coverage rate only means that there are some possibilities that the method would recover the truth, but it cannot guarantee that the method must obtain true matching. In other words, the coverage status $C_\text{BMCF}(t)$ of the bipartite method is a lower bound for the Pair Identity $\PI_\delta(t)$ of the reduced space method with parameter $\delta$, while the truth coverage status $C_\delta(t)$ of the reduced space method gives an upper bound, i.e.,
\begin{equation}
C_\text{BMCF}(t) \le \PI_\delta(t) \le C_\delta(t)\,.\label{eq:coverage_ineq}
\end{equation}
Hence the average truth coverage rate is an optimistic estimate for the pair identity and can be used to measure the (potential) matching performance.

\subsection{Results}

We conduct 100 independent experiments under each different setting that the expected number of cells $N_0$ goes from 15 to 50 with step 5, and the standard deviation $\sigma$ goes from 1 to 4.

\subsubsection{Tracking Performance}

\begin{figure}[H]
  \centering
  \includegraphics[width=\textwidth]{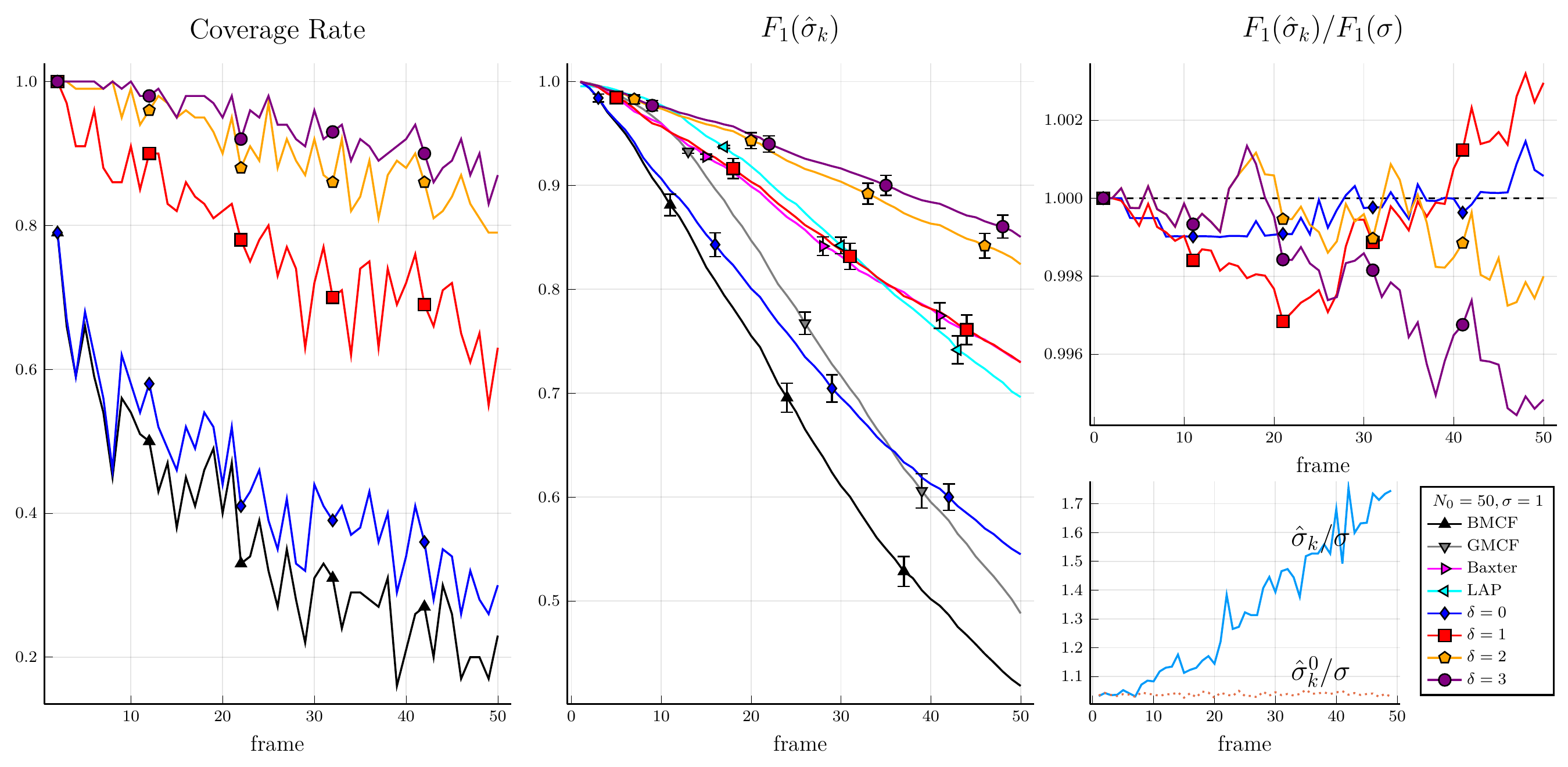}
  \caption{Performance of different methods, represented by distinct colored marker shapes shown in the bottom right legend, on 100 experiments under the simulation setting, $N_0=50, \sigma=1$. The left panel shows the average coverage rate of different proposed methods with the bipartite method BMCF in \textcite{padfieldCoupledMinimumcostFlow2011}, and the middle panel compares their cumulative accuracy with \textcite{jaqamanRobustSingleparticleTracking2008}'s two stages algorithm LAP and two global association methods, GMCF in \textcite{zhangGlobalDataAssociation2008} and Baxter in \textcite{magnussonGlobalLinkingCell2015}, where the error bars indicate 1.96 standard deviations. The upper right panel compares the cumulative accuracy $F_1(\hat\sigma_k)$ based on the estimated $\hat\sigma_k$ with the cumulative accuracy $F_1(\sigma)$ based on the oracle $\sigma$ by calculating their ratio. Furthermore, the bottom right panel checks the consistency of the estimation of $\sigma$, where the estimation $\hat\sigma_k^0$ given the real matching vectors is consistent, as shown in the dotted curve, but the solid curve indicates that the practical estimation $\hat\sigma_k$ would tend to increase along with the time frame.}
  \label{fig:res50_1}
\end{figure}

Figure \ref{fig:res50_1} summarizes the performance under the setting $N_0=50, \sigma=1$. Specifically, the left panel shows the coverage rate for different reduced space methods and the bipartite method (here BMCF), which always has a worse coverage rate than the proposed reduced methods, supporting the claim that the bipartite method serves as a lower bound \eqref{eq:coverage_ineq}. It is also reasonable to see that greater $\delta$ would have a higher coverage rate due to larger search spaces. Compared to the bipartite method, all reduced space methods can improve the coverage rate. In particular, the smallest nonzero $\delta=1$ substantially elevates the coverage rate at each frame, as shown by the largest margin between the curve $\delta=1$ and the curve $\delta=0$, which also brings non-neglected improvement. Largest $\delta=3$ can even do better, where the coverage rate at the last frame can be raised from 0.2 to nearly 0.9, although $\delta=1$ has elevated it to 0.6. These significant improvements show the reduced space strategy's efficiency since the much smaller search space taken from the original huge space can cover most and even nearly $100\%$ truth.

In addition to showing the efficiency of reduced space strategy, the coverage rate can also be viewed as the optimistic estimate, i.e., an upper bound, of the (pair identity) accuracy, but there is no clear relationship between the coverage rate and the accuracy. Would the method with a higher coverage rate also have a higher accuracy? The cumulative path accuracy $F_1(\hat\sigma_k)$ in the middle panel of Figure \ref{fig:res50_1} gives an answer, which shows that the bipartite method also serves as a lower bound in terms of the path accuracy, and a larger $\delta$ indeed gets a higher cumulative accuracy, not just increases the upper bound of pair identity. Note that here $\hat\sigma_k$ means that we estimate the velocity variance without assuming they are homogeneous among all frames, although the data generation scheme shares a common $\sigma$. Other approaches do not have $\sigma$ or equivalent parameters, but they have many other tuning parameters. We have conducted pre-experiments to optimize these tuning parameters to our best effort and refer to the Supplementary Material for more details. According to the up and down positions of these accuracy curves, we roughly have
$$
\{\delta=3\} > \{\delta=2\} > \{\delta=1\} \approx \text{Baxter}\approx \text{LAP} > \{\delta=0\} \approx \text{GMCF} > \text{BMCF}\,.
$$
The above rank is overall in terms of the path accuracy for the whole video, i.e., the rightmost points on the curves. Given a particular length, the performance might rank slightly differently, especially GMCF and LAP. With fewer frames, GMCF is as good as the reduced methods $\delta=2,3$, but it decreases sharply, and finally, even gets worse than $\delta=0$. LAP exhibits a similar pattern in the first few frames, but the slope of decreasing is much smaller than GMCF, and finally, it performs roughly equally well as $\delta=1$, although slightly worse.
The error bars indicate 1.96 standard deviations based on 100 multiple experiments, and these deviations tend to increase along with the frame, which means that some experiments might have much better accuracy than the mean accuracy curve, while some other experiments might have quite worse results than the mean curve.

\begin{figure}[H]
  \centering
  \includegraphics[width=\textwidth]{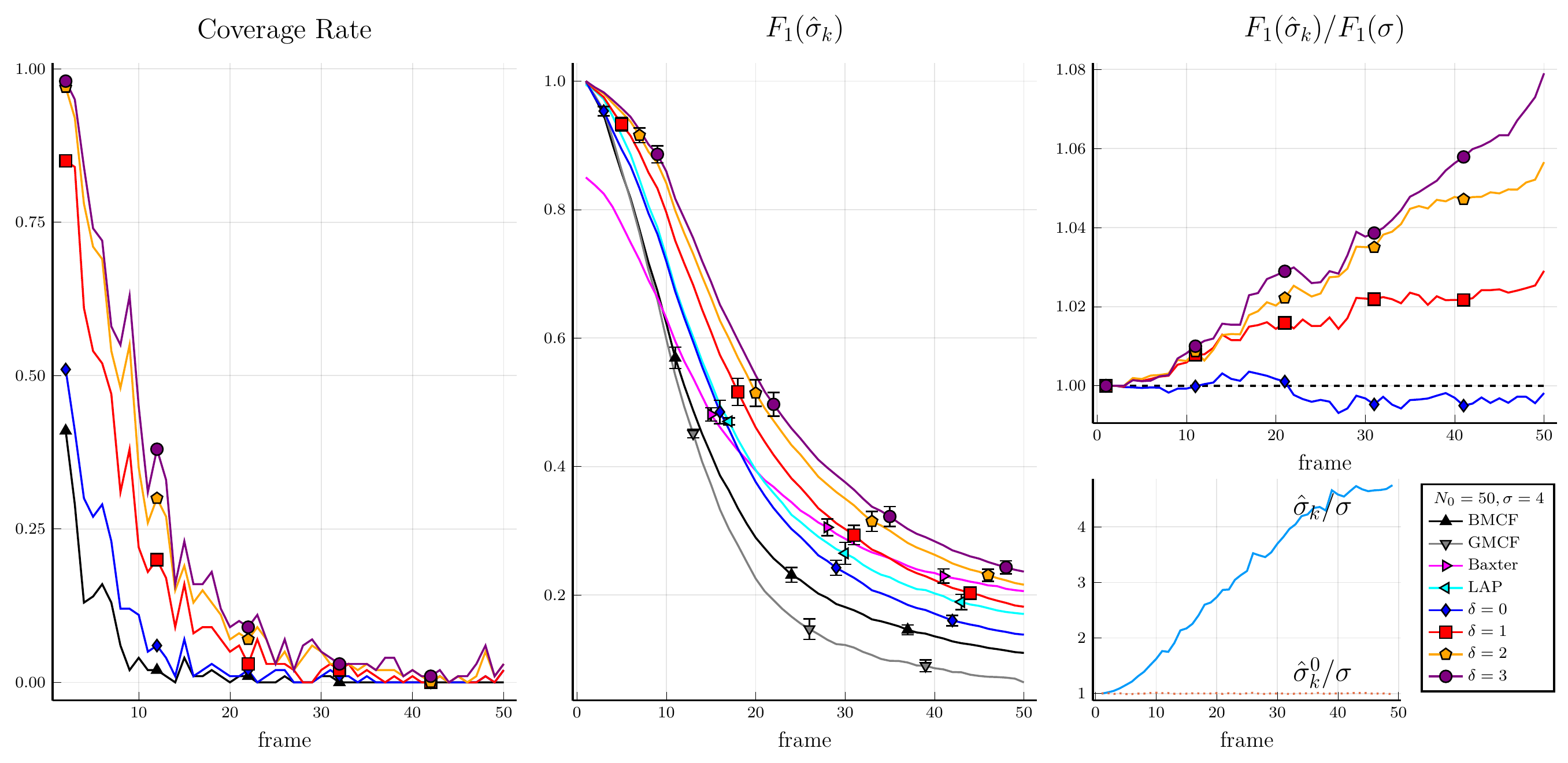}
  \caption{Similar to Figure \ref{fig:res50_1} except for different simulation setting $N_0=50, \sigma=4$, the performance of different methods on 100 experiments are summarized by the average coverage rate (left panel), the cumulative accuracy (middle panel), and the ratio of cumulative accuracy (upper right panel), as well as the ratio of estimation $\hat\sigma_k$ (bottom right panel).}
  \label{fig:res50_4}
\end{figure}

\begin{figure}[H]
  \centering
  \includegraphics[width=\textwidth]{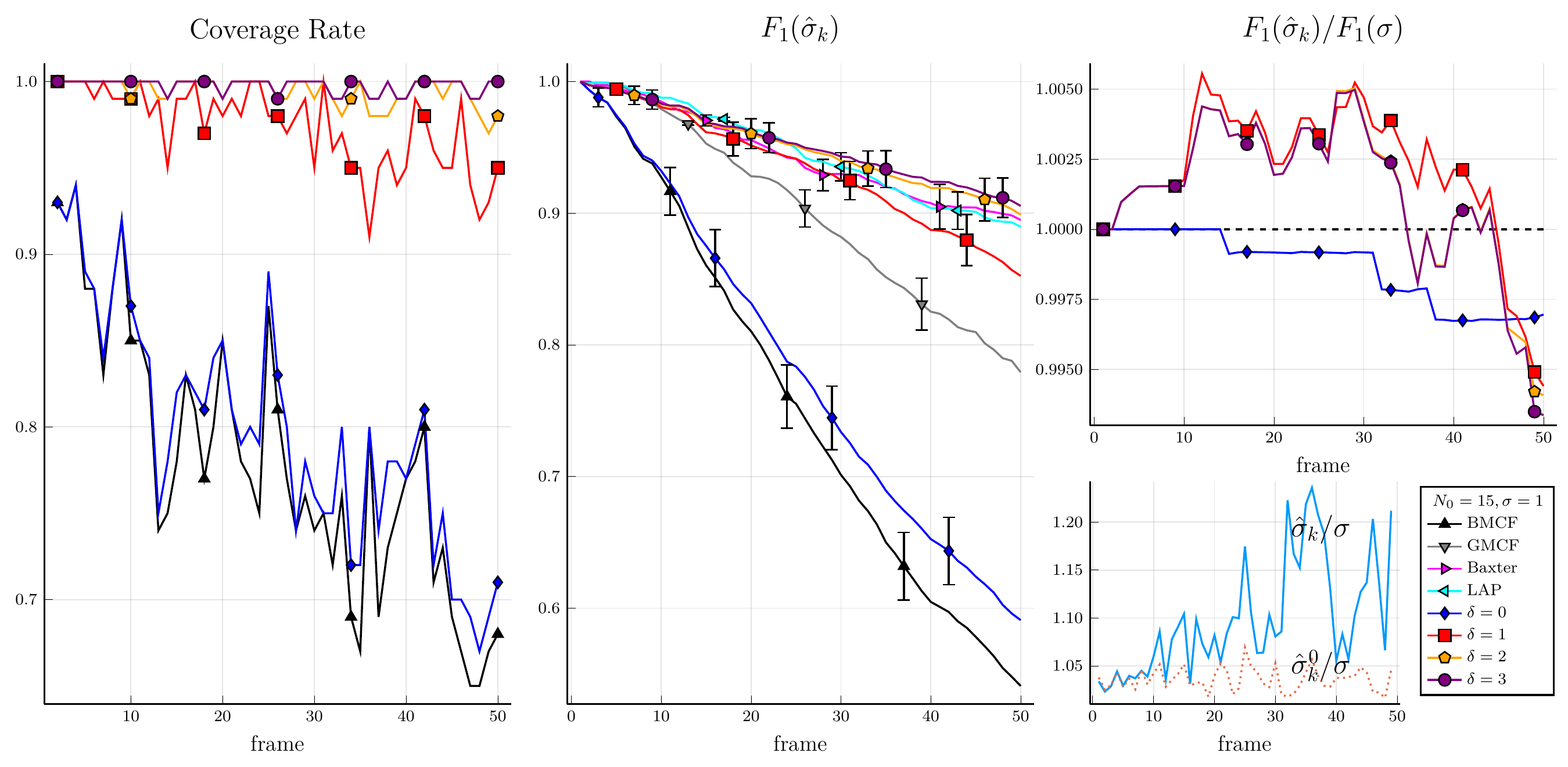}
  \caption{Summarize the performance of different methods on 100 experiments under the simulation setting $N_0=15, \sigma=1$ from the same four aspects used in Figure \ref{fig:res50_1} and Figure \ref{fig:res50_4}, which are the average coverage rate (left panel), the cumulative accuracy (middle panel), the ratio of cumulative accuracy (upper right panel), and the ratio of estimation $\hat\sigma_k$ (bottom right panel).}
  \label{fig:res15_1}
\end{figure}

To quantify the effect the parameter $\sigma$ on the matching performance, we calculate the ratio of the accuracy based on the estimated $\hat\sigma_k$ and the authentic $\sigma$, $F_1(\hat\sigma_k)/F_1(\sigma)$, for different reduced space methods, as shown in the top-right of Figure \ref{fig:res50_1}. Besides, the ratio $\hat\sigma_k/\sigma$ is presented in the bottom-right panel, as well as the ratio $\hat\sigma_k^0/\sigma$, where $\hat\sigma_k^0$ is the estimation given the real matching vectors. The ratio $\hat\sigma_k^0/\sigma$ fluctuates slightly around 1, which implies that the natural estimator proposed in Figure \ref{fig:estsigma} is reasonable, but $\hat\sigma_k$ tends to increase along with the time frame, which has been explained in Section \ref{sec:sigma}. The $F_1$ scores ratio shows that the matching under the estimated $\hat\sigma_k$ is not necessarily worse than the matching under the authentic $\sigma$. However, the bias from $F_1(\sigma)$, i.e., away from the line $y=1$, tends to increase, although with some turbulence. Similar trends can be found in other simulation settings, as shown in Figure \ref{fig:res50_4} and Figure \ref{fig:res15_1}. However, if we pay attention to the ticks on the $y$-axis, the maximum drift range is around (-0.006, +0.003), which implies that the effect of $\hat\sigma_k$ on the matching performance is quite minimal, if not negligible.

With increasing $\sigma$, the motions of objects would be faster and more variable, where the velocity might sometimes be somewhat fast and sometimes relatively slow, and even suddenly be in the opposite direction, although a smaller $\sigma$ also allows direction changes but with a narrower range. So it would be more challenging to get good matching results,  just as the worse coverage rate and cumulative accuracy are shown in Figure \ref{fig:res50_4} for the setting $N_0=50, \sigma=4$. The coverage rate for all methods decreases sharply, and the bipartite method even drops to zero, which means that no experiment among 100 experiments obtains the real pairwise matching vector. The cumulative accuracy also decreases quickly, and the whole path accuracy of the distance method is around 0.1, much less than 0.4 in Figure \ref{fig:res50_1}. The reduced space methods also exhibit much worse performance, although they, in particular $\delta=2,3$, are still better than other methods, and larger $\delta$ again performs better. GMCF seems more sensitive to the variable motions, which can get better results than BMCF when $\sigma=1$, and even outperform our proposed $\delta=1$ in the first half part frames, but it always stays as the worst method when $\sigma=4$ in all frames. The order of path accuracy would stay the same,
$$
\{\delta=3\} > \{\delta=2\} > \{\delta=1\} > \text{LAP} > \{\delta=0\} > \text{BMCF} > \text{GMCF}\,,
$$
along with the time frame except for Baxter, which shows more robust to the length of the trajectories. Although Baxter's path accuracy for the first several frames is the worst, it exceeds other methods' performance successively with a much lower decreasing rate and finally becomes better than $\delta=1$. As for the ratio of $F_1$ scores, the maximum drift is around 0.08, which means that the effect of $\hat\sigma_k$ is moderate, although it is larger than the maximum drift when $\sigma=1$, and smaller $\delta$ tends to be less inconsistent.

The performance would be much better if the expected number of cells decreased to $N_0=15$, as shown in Figure \ref{fig:res15_1}. The worst coverage rate is around 0.7, much better than the setting $N_0=50$, and the best coverage rate even always stays at 1.0, which means that we could obtain the entirely correct matching results. The cumulative accuracy curves for the reduced methods $\delta=2,3$ again keep the top two, followed by LAP and Baxter, then next $\delta=1$ beats GMCF,  and $\delta=0$ defeats BMCF, that is
$$
\{\delta=3\} \approx \{\delta=2\} > \text{LAP} \approx \text{Baxter} > \{\delta=1\} > \text{GMCF} > \{\delta=0\}  > \text{BMCF}\,.
$$
The close gap between the top three methods also conveys the message that taking $\delta=2$ already has a significant improvement, and there might not be necessary to get a further improvement with some additional computational cost. Besides, the estimation of $\hat\sigma_k$ seems much more consistent than other settings, and the maximum drift of the $F_1$ score is also relatively minimal.

\subsubsection{Choice of $\delta$}

Slightly abusing the notation, let $F_1(\delta)$ be the whole path accuracy for the reduced method with parameter $\delta$, and $F_1(-1)$ denotes the accuracy for the bipartite method, BMCF. From Figure \ref{fig:res50_1}, \ref{fig:res50_4}, \ref{fig:res15_1}, a larger $\delta$ always brings the most substantial accuracy improvement $F_1(\delta)-F_1(-1)$ over the bipartite method, but it does not mean that the largest $\delta$ would be the best choice in practice since there is a tradeoff between the computational cost and accuracy. Sometimes the performance is good enough without the necessity to increase $\delta$, such as the tight top three cumulative accuracy curves in Figure \ref{fig:res15_1}. Consider the relative amount of improvements for the reduced methods $F_1(\delta)-F_1(\delta-1)$ in these three figures. The gap achieves the largest when $\delta=1$. Moreover, we have checked that all conducted simulations show the same phenomenon, that the amount of accuracy improvement between $\delta=0$ and $\delta=1$ is the most substantial.

\begin{figure}[H]
  \centering
  \begin{subfigure}{0.5\textwidth}
    \centering
    \includegraphics[width=\textwidth]{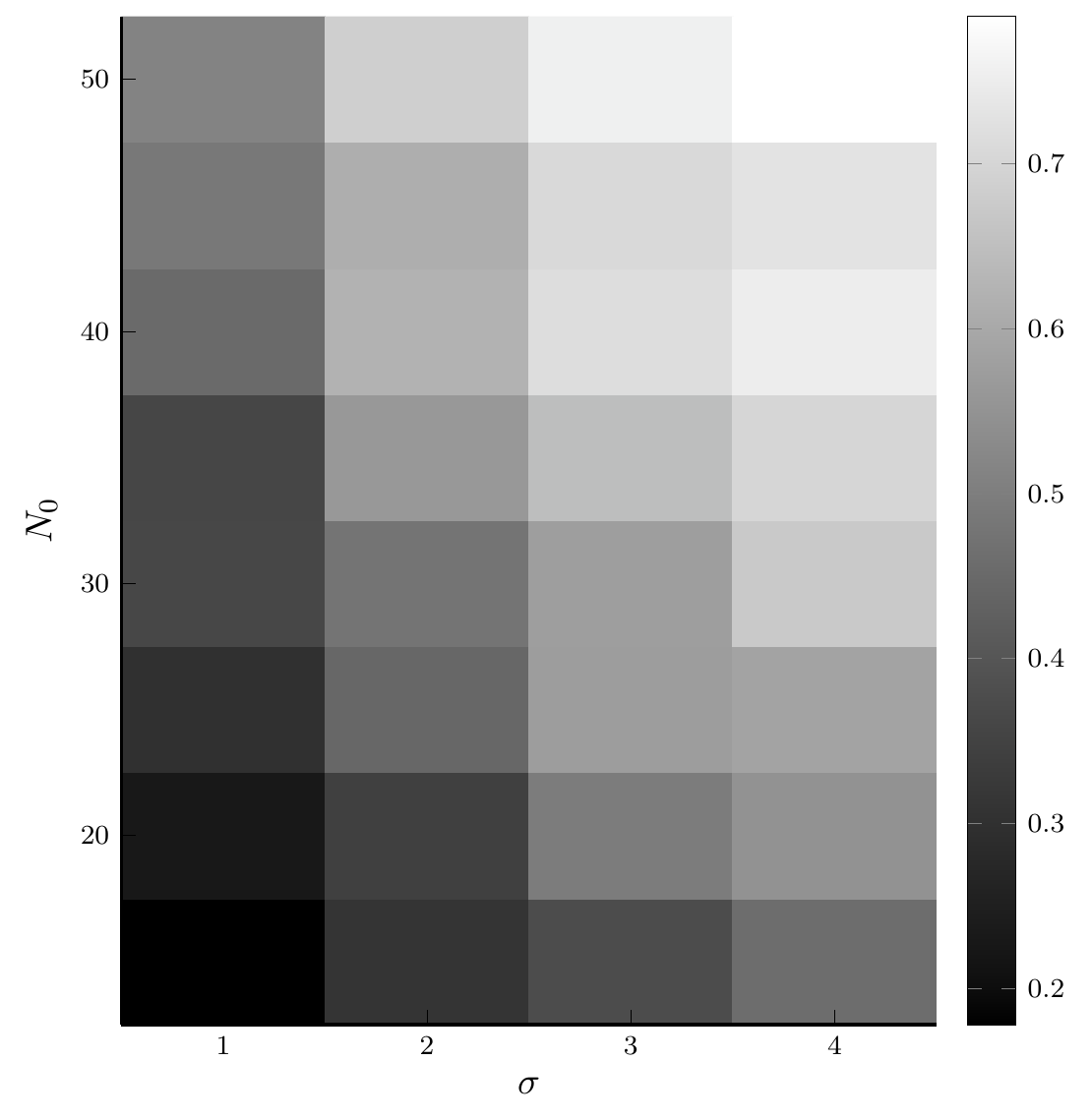}
    \caption{Whole path accuracy.}
    \label{fig:wpa_delta1}
  \end{subfigure}%
  \begin{subfigure}{0.5\textwidth}
    \centering
    \includegraphics[width=\textwidth]{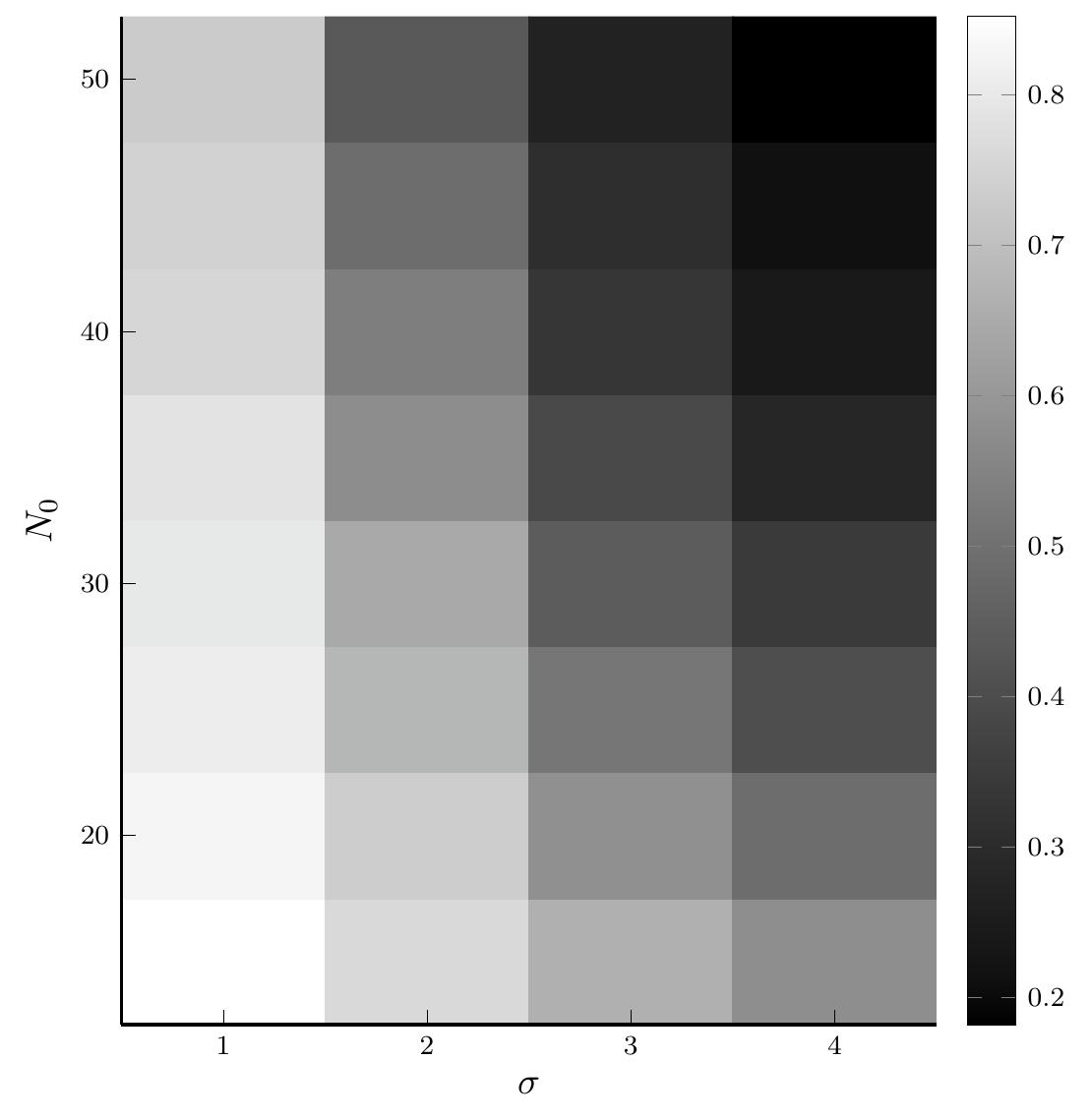}
    \caption{Ratio of relative accuracy improvement.}
    \label{fig:ratio_improve}
  \end{subfigure}
  \caption{The performance of the reduced space method with $\delta=1$ under all simulation settings, where $N_0$ starts from 15 to 50 with step 5, and $\sigma$ varies from 1 to 4. The small rectangles in these two heatmaps represent the whole path accuracy $F_1(\delta = 1)$ and the ratio of relative accuracy improvement $\frac{F_1(2) - F_1(1)}{F_1(1) - F_1(0)}$, respectively. The brighter color represents a higher value, as illustrated in the color bar on the right-hand side.}
  \label{fig:compare_delta1}
\end{figure}

To investigate the performance of our proposed method under different simulation settings, Figure \ref{fig:wpa_delta1} shows the whole path accuracy for $\delta=1$ under different $N_0$ and $\sigma$, where brighter color represents higher accuracy, which implies that higher $N_0$ and higher $\sigma$ tend to worse matching performance. The worst accuracy would be only around 0.2, corresponding Figure \ref{fig:res50_4}, so the choice of $\delta=1$ seems not enough, although it is sufficient in Figure \ref{fig:res15_1} since the accuracy is already improved to be around 0.85 and litter improvement if we continue to increase $\delta$.
Consider the ratio of the relative accuracy improvement,
$$
R(\delta) = \frac{F_1(\delta+1) - F_1(\delta)}{F_1(\delta) - F_1(\delta-1)}\,,\qquad \delta=0,1,2\,.
$$
Figure \ref{fig:ratio_improve} displays $R(1)$ by the nearly opposite heatmap of Figure \ref{fig:wpa_delta1}, and it suggests that the simulations with larger $N_0$ and larger $\sigma$ should increase $\delta$ to get better performance since the higher ratio implies that there is still substantial improvement can be obtained.

\subsubsection{Computational Speed}

The corresponding average computation times of the experiments have been summarized in Table \ref{tab:runtime}. Generally, the running time will increase with the number of cells regardless of the methods. For a moderate number of cells ($N_0=15$), our proposed approaches can be as fast as others, and even faster than the Baxter algorithm, and it is also bearable as an offline algorithm for a large number of cells ($N_0=50$).

\begin{table}[H]
  \centering
  \input{tabs/runtime.tex}
  \caption{Computational speed of different methods. The first two columns indicate the experiment setting on the expected number of cells $N_0$ and the variance level $\sigma$. The following columns, except for the rightmost three columns containing the ratios $r_{\delta,\delta-1}\,,\delta=1,2,3$, are the average running time measured in the 100 experiments, where the values are in seconds for $N_0=15$, and in minutes for $N_0=50$.}
  \label{tab:runtime}
\end{table}

Based on the computation time, we try to validate the complexity analyzed in Proposition \ref{prop:reduced_complexity}, which claims that the complexity is $O((\delta+1/2)^2N^4)$, where $N$ roughly equals $N_0$. We estimate the complexity by the observed running time and let
$$
r_{ij} = \frac{\text{running time of reduced method with }\delta=i}{\text{running time of reduced method with }\delta=j}\,,
$$
while a natural approximation for the theoretical ratio is
$$
r_{ij}^\star = \left(\frac{i + 1/2}{j + 1/2}\right)^2\,.
$$
Since higher memory requirement usually slows down the speed, it would be more proper to compare the complexity given the same memory allocation. In our experiments, the memory allocation is dominated by the size of search space, and larger $\delta$ would require larger memory. There is little (although not no) difference between the memory requirement by two consecutive $\delta$'s, so we consider the ratio between two consecutive $\delta$'s instead of the ratio like $r_{30}$ to alleviate the side effect of memory allocations. The rightmost three columns in Table \ref{tab:runtime} present the observed ratios $r_{\delta,\delta-1}, \delta=1,2,3$, where the observed ratio is quite close to the theoretical ratio. The ratios exhibit an increasing pattern along $\sigma$, which could be explained by more disappearing cells in the simulation with a larger $\sigma$ and hence less accurate in the bound of complexity (see the proof of Proposition \ref{prop:reduced_complexity}).

\begin{figure}[H]
  \centering
  \includegraphics[width=\textwidth]{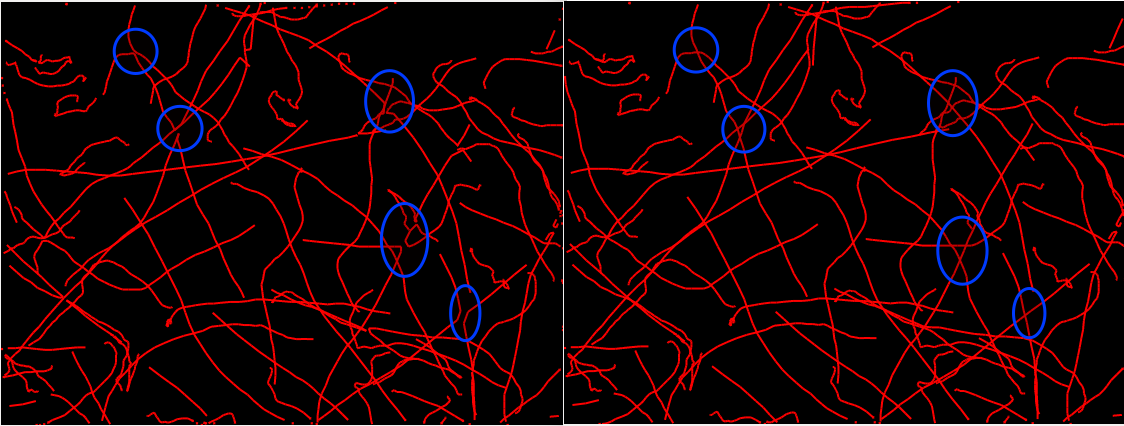}
  \caption{In a simulation under setting $N_0=50,\sigma=1$, all paths obtained by the bipartite method (left panel) and the proposed method with $\delta=1$ (right panel).}
  \label{fig:demo_sim}
\end{figure}

To check how we can do better with the proposed methods, we pick one experiment under the setting $N_0=50, \sigma=1$ as an example. Figure \ref{fig:demo_sim} compares the trajectories obtained by the bipartite method (left panel) and our proposed method with $\delta=1$ (right panel). Each red curve represents a path, and the blue ellipses mark the differences obtained by these two methods. All the true paths in the ellipses regions agree with those obtained by the proposed method and exhibit a cross-path pattern, in which the bipartite method would always fail, as discussed in Proposition \ref{prop:cross}.

\section{Cell Tracking Challenge}\label{sec:ctc}

To further demonstrate the performance, we compare our approach with the methods which achieve outstanding accuracy on some datasets in the Cell Tracking Challenge (CTC). Here we choose datasets \datasim\ and \datagowt\ since the cells look similar to those in the video we considered in Section \ref{sec:real-cell-video}, i.e., nearly circular and roughly the same size.

\begin{figure}[H]
  \centering
  \begin{subfigure}{0.495\textwidth}
    \centering
    \includegraphics[width=\textwidth, height=\textwidth]{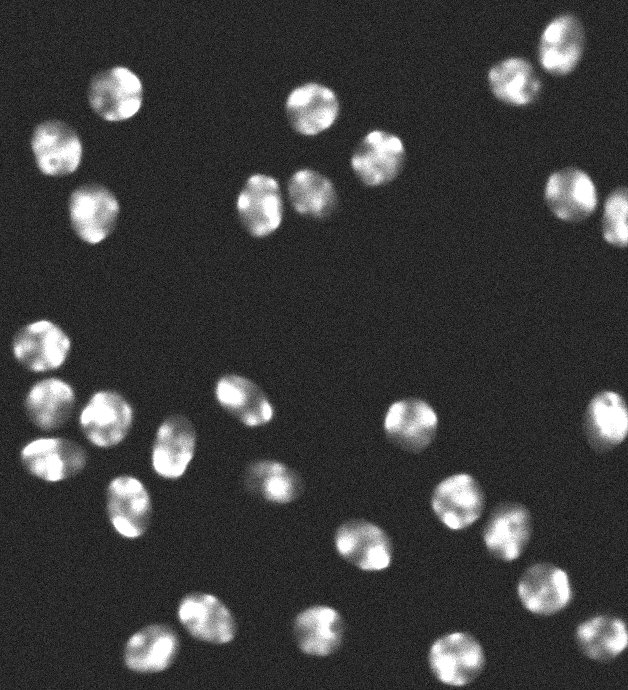}
    \caption{First frame of \datasim.}
  \end{subfigure}%
  \begin{subfigure}{0.495\textwidth}
    \centering
    \includegraphics[width=\textwidth, height=\textwidth]{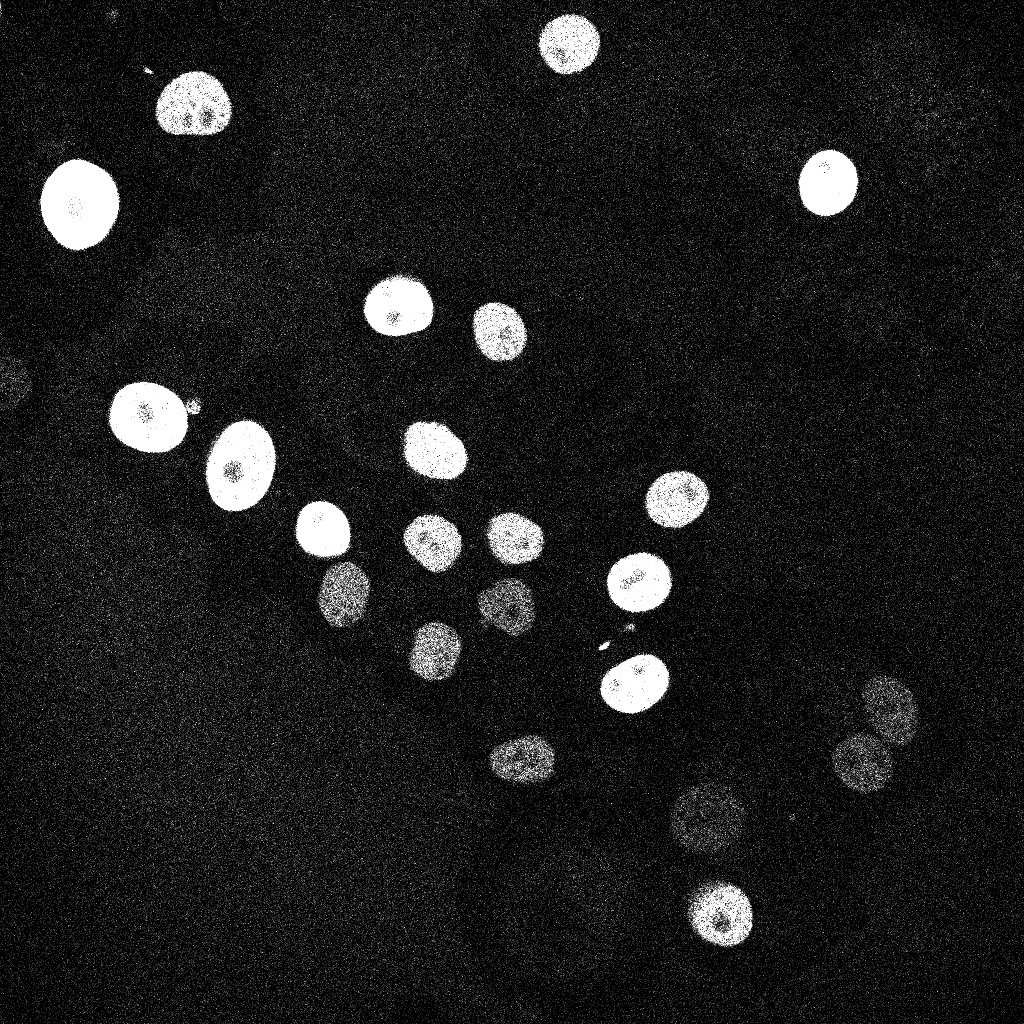}
    \caption{First frame of \datagowt.}
  \end{subfigure}
  \caption{Sample images from two datasets. The brightness and contrast have been adjusted for legibility.}
\end{figure}

We investigate two top methods with the following performances (as of 2021-05-09).

\begin{itemize}
  \item TUG-AT \parencite{payerSegmentingTrackingCell2019}: the tracking measurement ranks 1/35 on \datagowt, and the overall performance (named \OPCTB) is 5/35, which takes the segmentation into account; on \datasim, the ranks are 2/33 and 12/33, respectively.
  \item KTH-SE \parencite{magnussonGlobalLinkingCell2015} (it is actually the Baxter algorithm discussed in Section \ref{sec:simulations}): the tracking measurement ranks 2/35 on \datagowt, and the overall performance is 1/35; on \datasim, the ranks are 10/33 and 9/33, respectively.
\end{itemize}

Note that both methods consist of segmentation and association, but our proposed approach focuses on the second step -- association; thus, direct comparisons with input as the raw image sequence would be unsuitable.
Alternatively, we can compare the tracking performance based on the segmentation results, as shown in the following workflow.

\begin{figure}[H]
  \centering
  \includegraphics[width=\textwidth]{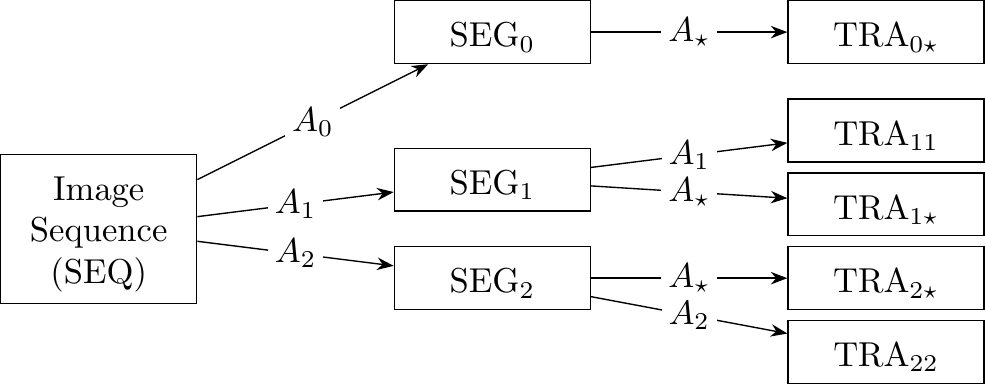}
  \caption{Workflow for comparing the proposed approach $A_\star$ with top approaches $A_1, A_2$ in CTC on an image sequence named SEQ. The segmentation $\SEG_i, i=0,1,2$ is obtained by approach $A_i$, where $\SEG_0$ is the ground truth by experts' labeling ($A_0$). The tracking result $\TRA_{ij}$ is obtained by conducting the tracking method $A_j, j=\star,1,2$ on the segmentation $\SEG_i$.}
  \label{fig:workflow}
\end{figure}

For each competing approach $A_i, i=1,2$, conduct it on an image sequence and obtain the final tracking result $\TRA_i$, together with the segmentation $\SEG_i$. Then perform the tripartite matching (and bipartite matching) $A_\star$ on the segmentations generated by the competitors. In addition, since the ground truth (GT) of segmentation is also available, we can evaluate our proposed approaches on the ground truth. For consistent comparisons, we adopt the Acyclic Oriented Graph Matching \parencite{matulaCellTrackingAccuracy2015} tracking measurement in CTC, which falls in $[0, 1]$ with higher values corresponding to better tracking performance.

Note that our approach assumes no splitting cells, and hence no cell appearing and disappearing from the middle. However, these two datasets allow the splitting behavior. For a more fair comparison, we divide the whole image sequence into several sub-sequences to eliminate the splitting behavior, i.e., cut the sequence at the images where there are splitting events. Then pick the sub-sequences with the number of images larger than some threshold, say 10. Table \ref{tab:ctc_sub} displays the results, where the columns $S$ and $T$ represent the starting index and the ending index of a sub-sequence.

Although the competing methods can be viewed as tracking followed by segmenting, in practice, it is cumbersome and possibly problematic to separate the whole program into the segmentation part and tracking part, so the competing approaches on others' segmentations, such as $A_1$ on $\SEG_2$, are inapplicable, and just leave them blank in the table.

\begin{table}[H]
  \centering
  \resizebox{\textwidth}{!}{%
    \input{tabs/subacc_ctc_v2.tex}  }
  \caption{Tracking accuracy following the workflow in Figure \ref{fig:workflow}. Three segmentation approaches (the ground truth by experts' labelling, $\SEG_1$ by KTH-SE, and $\SEG_2$ by TUG-AT) are performed on the sub-video from frame $S$ to frame $T$ in each sequence SEQ of two datasets. Then the tracking approaches (two CTC competitors $A_i,i=1,2$, bipartite matching BMCF, tripartite matching $\delta=0,1,2,3$) are conducted on the segmentations. Finally, the tracking results $\TRA$ are compared with the underlying truth to calculate the accuracies. The blanks imply that the corresponding results are not applicable.}
  \label{tab:ctc_sub}
\end{table}

Table \ref{tab:ctc_sub} shows that our approaches can achieve the same accuracy in most sub-sequences, and sometimes even better, such as the sub-sequence [29, 38] of sequence 02 based on TUG-AT's segmentation and sub-sequence [37, 48] of sequence 02 based on KTH-SE's segmentation. Moreover, the tracking accuracies based on the inferred segmentations are pretty close to the ones based on ground truth segmentation, indicating that the effect of the segmentation error is low. Hence, it validates our argument that the segmentation can be assumed to be accurate enough.

Overall, we conclude that the proposed approaches can achieve comparable performance as the top methods in the CTC. Refer to Supplementary Material for more details on the implementations, together with the comparison results on the whole sequences.

\section{Real Cell Video}\label{sec:real-cell-video}

Primary human NK cells were isolated from fresh PBMCs by negative selection using the EasySep Human NK Cell Enrichment Kit (Stemcell), according to the manufacturer's protocol. The NK cells were then co-cultured with a human cancer cell line, U-2 OS, in phenol red-free CO2-independent medium (Invitrogen) supplemented with 10\% heat-inactivated FCS, 50 ng/ml IL-2, 100 U/ml penicillin, and 100 ug/ml streptomycin.  Cell images were acquired by phase-contrast imaging using a Nikon TE2000-PFS inverted microscope enclosed in a humidified chamber maintained at 37\textdegree C. Cells were imaged every 30 seconds by a motorized stage and a 20X objective (NA=0.95).

Figure \ref{fig:realall} shows the results of the 30 frames by the distance-based bipartite matching, denoted by blue curves, and our proposed tripartite matching, represented by orange curves. Each curve represents a cell trajectory, which starts at a particular frame and ends at another frame. In most regions, these two different colored curves coincide, which means the matching results by two different methods are the same, but there are still some diverged curves, such as the paths in the region marked by the red ellipses.

\begin{figure}[H]
\centering
\begin{subfigure}{\textwidth}
  \centering
  \includegraphics[width=\textwidth]{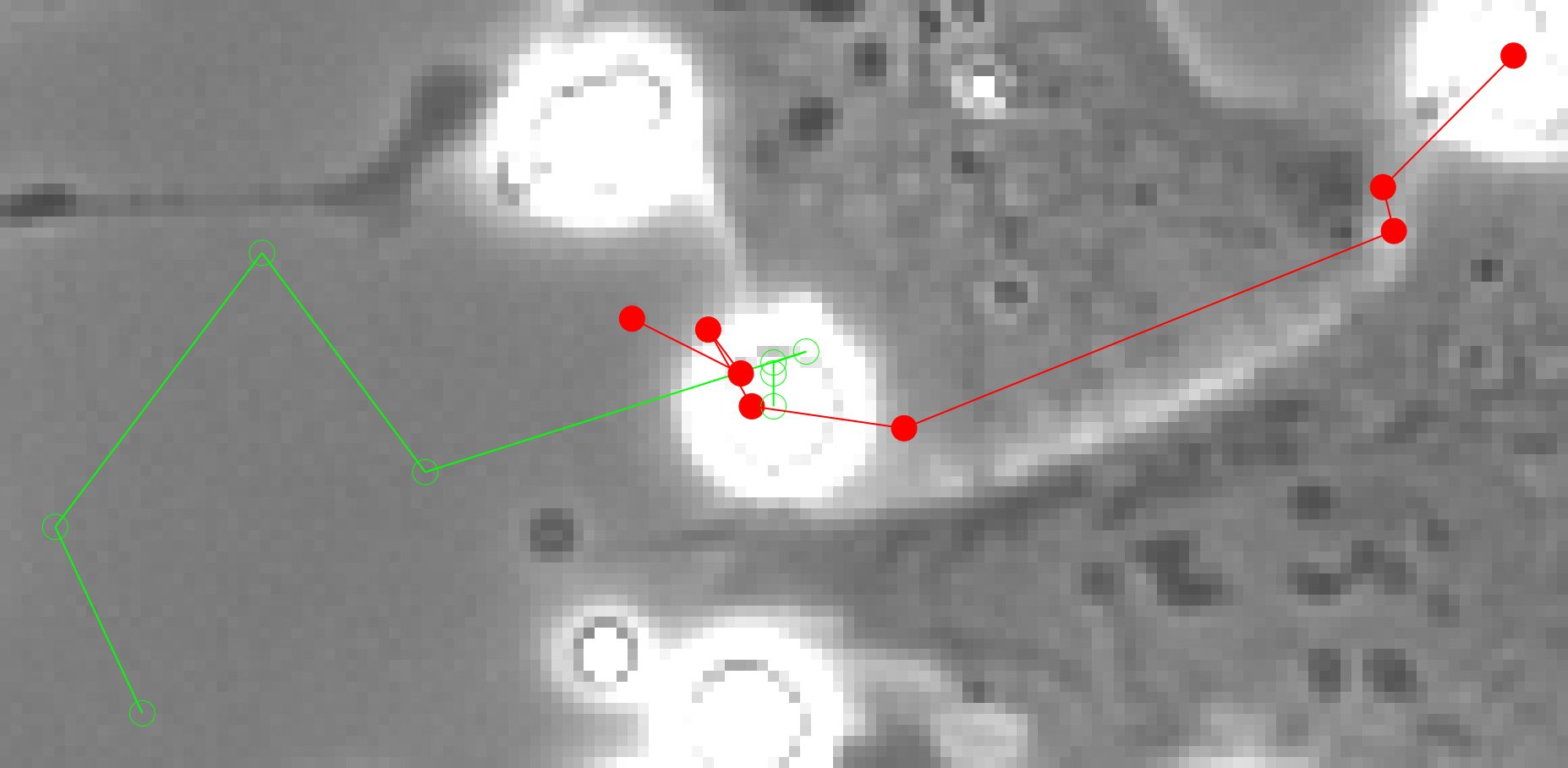}
  \caption{Bipartite matching.}
\end{subfigure}

\begin{subfigure}{\textwidth}
  \centering
  \includegraphics[width=\textwidth]{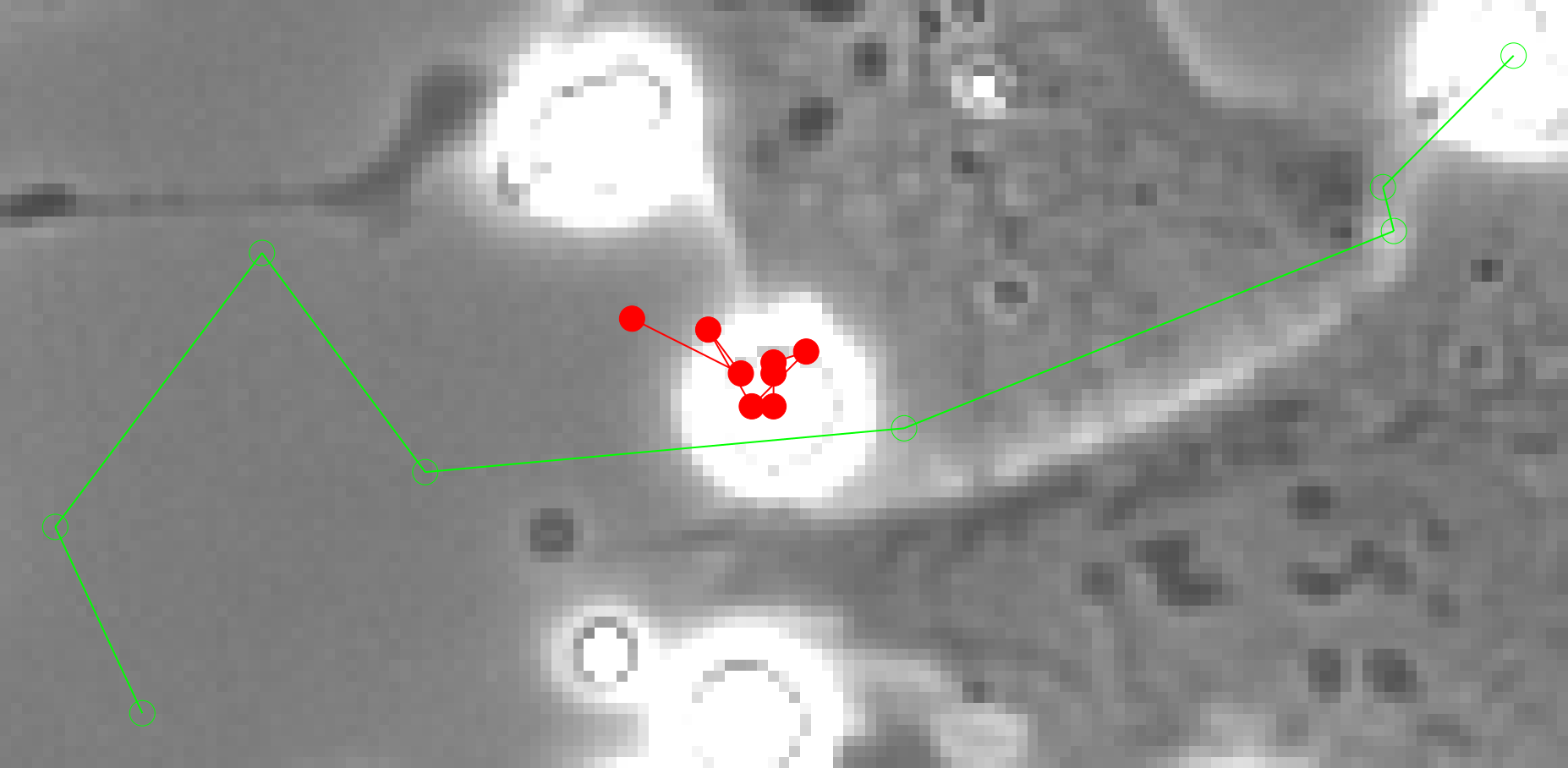}
  \caption{Tripartite matching.}
  \label{fig:real21to28b}
\end{subfigure}
\caption{Two sub-paths obtained from the bipartite matching (top panel) and the tripartite matching (bottom panel) in the red rectangle region of Figure \ref{fig:realall}.}
\label{fig:real21to28}
\end{figure}

Figure \ref{fig:real21to28} zooms into such a region, and it displays two paths obtained from the bipartite matching (top panel) and tripartite matching (bottom panel). We pick two cells, represented by the green and red circle, respectively, with their associated sub-paths. The background corresponds to the last frame of the sub-paths, i.e., the green (or red) circle that coincides with the real NK cell is the endpoint of its corresponding path, and then another end of the sub-paths represents the starting point. By careful observation from the raw video, we prefer to take the paths obtained from the tripartite matching, shown in Figure \ref{fig:real21to28b}, as the actual paths, where the hollow green cell moves faster than the solid red cell, and the hollow green cell has a clear direction while the solid red cell somewhat walks randomly. In contrast, the bipartite matching makes mistakes when the hollow green cell passes by the solid red cell. It forces the hollow green cell to slow down suddenly and even be still but lets the solid red cell become directional and speed up quickly, both of which are somewhat unrealistic.

\begin{figure}[H]
  \centering
  \begin{subfigure}{0.5\textwidth}
    \centering
    \includegraphics[width=0.99\textwidth]{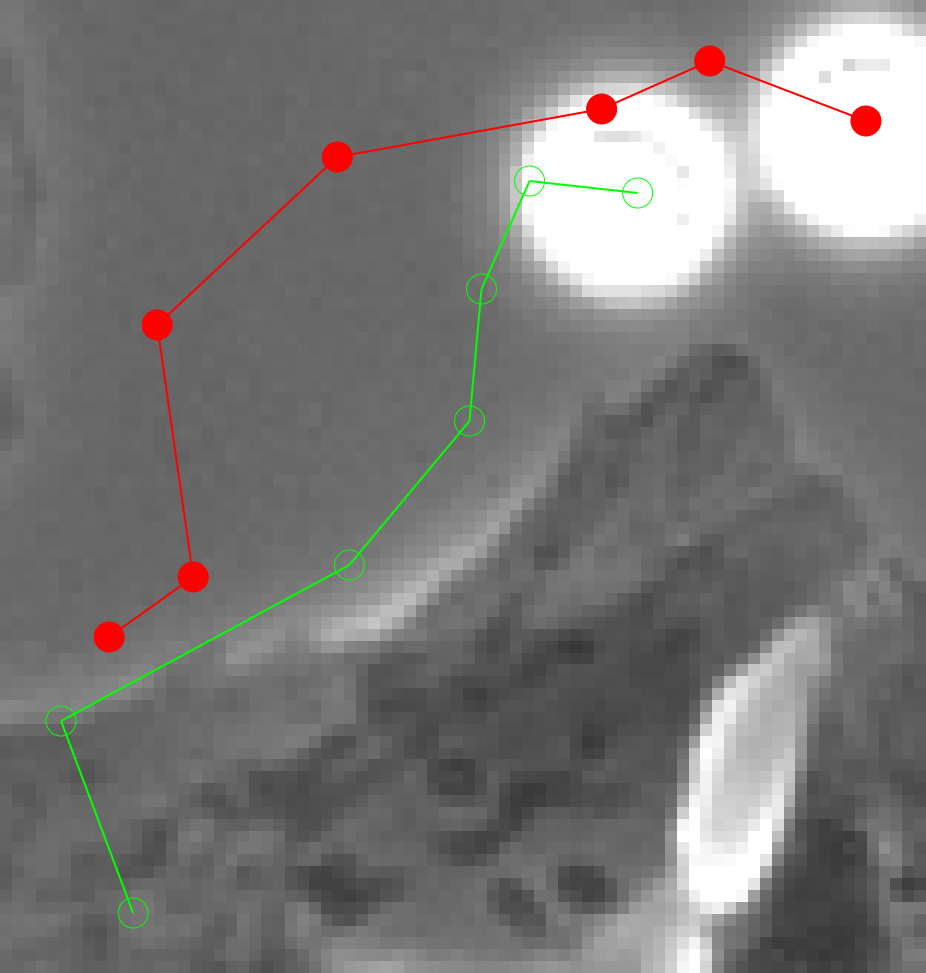}
    \caption{Bipartite matching.}
    \label{fig:real3to9a}
  \end{subfigure}%
  \begin{subfigure}{0.5\textwidth}
    \centering
    \includegraphics[width=0.99\textwidth]{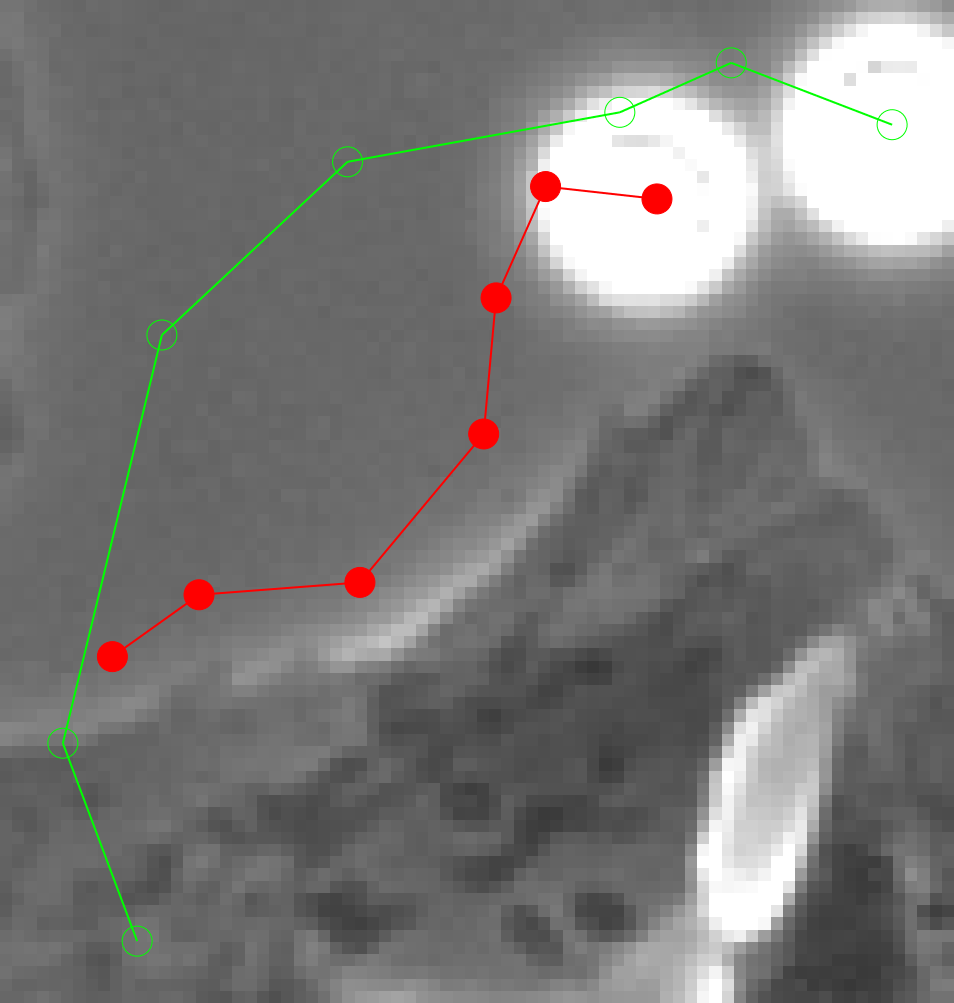}
    \caption{Tripartite matching.}
    \label{fig:real3to9b}
  \end{subfigure}
  \caption{Similar to Figure \ref{fig:real21to28} in the red rectangle region of Figure \ref{fig:realall}, but two different sub-paths obtained from the bipartite method (left panel) and the tripartite method (right panel).}
  \label{fig:real3to9}
\end{figure}

There is another pair of different paths in the red rectangle box of Figure \ref{fig:realall}, and shown in Figure \ref{fig:real3to9}. Again with careful observation, we prefer the paths in Figure \ref{fig:real3to9b} obtained by the tripartite matching, where the hollow green cell has a higher speed, and both cells change their direction steadily. However, the bipartite method suddenly alters the directions when matching the second and third frames, as shown in Figure \ref{fig:real3to9a}. There is a cross-path pattern, where the hollow green cell should move upward, and the solid red cell moves to the right, that the bipartite method would always fail, which forces the hollow green cell to suddenly turn right and lets the solid red cell move upward immediately.

Both Figure \ref{fig:real21to28} and \ref{fig:real3to9} are examples of Proposition \ref{prop:cross}, where the bipartite matching fails in these cross-path situations. In contrast, our proposed tripartite matching explores other matching vectors, whose spaces are constructed by exchanging the bipartite matching result, to minimize our defined goal function that prefers smooth velocity changes. Thus, it can correct the erroneous paths generated by the traditional bipartite matching.

\section{Conclusion}\label{sec:conclusion}

We have presented a tripartite matching framework for multiple object tracking. Contrary to many tracking methods developed for particular objects by appearance modeling, we aim to forgo the appearance and instead model the pure movement for some appearance-free tracking tasks. We formulate the tripartite matching as maximizing the likelihood of the state vectors constituted by the position and velocity and employ the dynamic programming to solve the maximum likelihood estimate. The matching vector constructs the search space for dynamic programming, which could be huge when there are many objects. To overcome the computational cost induced by the large search space, we decomposed such space by the number of disappearing cells and proposed the reduced-space approach by truncating the decomposition. The investigation on the solution space helps perform a more organized and comprehensive searching than the existing velocity-based methods to avoid truncating high-quality parts and avoid trapping into local modes.

Here are some limitations of our proposed method. We truncate the search space to allow only one pair to exchange, which might be not enough, and that might be one reason for the worse performance in the larger $\sigma$ situations. If the user can bear higher computing burden, it is straightforward to modify our algorithm to allow more pairs to exchange. Although the estimation of $\Sigma_k$ has limited impacts on the matching performance, it tends to become more and more inconsistent, and hence it would be better to propose some less inconsistent (or even consistent) estimator.

Although the NK cells often move freely and smoothly, the bumping on the cancer cell or the collisions with other NK cells might violate the smoothing velocity assumption. In these situations, the proposed tripartite method might be worse than the bipartite approach. Hopefully, the collisions between two moving NK cells are quite rare when the moving distances are much larger than the cell size. Besides, the bumping of NK cells on the cancer cell scarcely significantly changes their directions, i.e., the NK cells tend to move along the edge of the cancer cell instead of bouncing back.
Nevertheless, it would be more sensible to design some comprehensive methods which can incorporate the collision cases, such as adaptively switching between the tripartite matching and bipartite matching to handle non-collision and collision cases.

The assumption that no objects disappear (appear) from the middle is reasonable in 2D when the objects are restricted to a plate, but in more real situations, objects move in 3D; thus, extending the tripartite approach by removing such an assumption would be attractive. Moreover, it is desirable to integrate the pure motion model with a dynamic appearance model to track objects' morphological changes.

\section*{Acknowledgment}
This research is supported by the Hong Kong PhD Fellowship Scheme from the University Grant Committee and two grants from the Research Grants Council (14303819, C2006-17E) of the Hong Kong SAR, China.

\printbibliography

\end{document}

%% file: tabs/runtime.tex
\begin{tabular}{ccccccccccccc}
\toprule
$N_0$ & $\sigma$ & BMCF & GMCF & Baxter & LAP & $\delta =0$ & $\delta =1$ & $\delta =2$ & $\delta =3$ & $r_{10}$ & $r_{21}$ & $r_{32}$\tabularnewline\midrule
\multirow{4}{1cm}{\centering 15 (sec)} & 1 & 1.52 & 11.39 & 82.80 & 8.39 & 11.18 & 16.15 & 33.18 & 52.91 & 1.44 & 2.05 & 1.59\tabularnewline
 & 2 & 1.51 & 11.21 & 82.76 & 8.19 & 10.67 & 16.47 & 33.98 & 50.96 & 1.54 & 2.06 & 1.50\tabularnewline
 & 3 & 1.51 & 10.93 & 82.54 & 8.22 & 10.17 & 14.44 & 30.98 & 48.17 & 1.42 & 2.14 & 1.56\tabularnewline
 & 4 & 1.51 & 10.81 & 82.40 & 8.47 & 10.03 & 14.35 & 31.98 & 49.43 & 1.43 & 2.23 & 1.55\tabularnewline
\midrule
\multirow{4}{1cm}{\centering 50 (min)} & 1 & 0.10 & 2.54 & 3.01 & 0.34 & 5.39 & 40.34 & 89.53 & 136.82 & 7.48 & 2.22 & 1.53\tabularnewline
 & 2 & 0.10 & 2.93 & 3.15 & 0.36 & 5.56 & 48.10 & 118.77 & 186.73 & 8.65 & 2.47 & 1.57\tabularnewline
 & 3 & 0.10 & 2.88 & 2.83 & 0.33 & 4.68 & 42.24 & 108.74 & 179.04 & 9.03 & 2.57 & 1.65\tabularnewline
 & 4 & 0.09 & 2.83 & 2.59 & 0.34 & 4.17 & 38.09 & 99.77 & 171.36 & 9.14 & 2.62 & 1.72\tabularnewline
\midrule
\multicolumn{10}{r}{approximated theoretical ratio (large $N$): }& 9.00& 2.78& 1.96\tabularnewline\bottomrule
\end{tabular}

%% file: tabs/subacc_ctc_v2.tex
\begin{tabular}{cclllllllll}
\toprule
\midrule
\multirow{2}{*}{SEQ} & \multirow{2}{*}{SEG}& \multirow{2}{*}{S} & \multirow{2}{*}{T}& \multicolumn{7}{c}{TRA}\tabularnewline
\cmidrule{5-11}
 & & & & BMCF & $\delta = 0$ & $\delta = 1$ & $\delta = 2$ & $\delta = 3$ & KTH-SE & TUG-AT\tabularnewline
\midrule
\midrule
\multicolumn{11}{l}{Dataset: Fluo-N2DH-SIM+}\tabularnewline
\midrule
\multirow{4}{*}{01} & \multirow{1}{*}{GT} & 0 & 9 & 1.0 & 1.0 & 1.0 & 1.0 & 1.0 &  & \tabularnewline
\cmidrule(lr){2-11}
 & \multirow{2}{*}{KTH-SE} & 0 & 9 & 1.0 & 1.0 & 1.0 & 1.0 & 1.0 & 1.0 & \tabularnewline
 &  & 54 & 64 & 0.981585 & 0.981585 & 0.981585 & 0.981585 & 0.981585 & 0.989199 & \tabularnewline
\cmidrule(lr){2-11}
 & \multirow{1}{*}{TUG-AT} & 0 & 10 & 0.997345 & 0.997345 & 0.997345 & 0.997345 & 0.997345 &  & 0.997345\tabularnewline
\midrule
\multirow{12}{*}{02} & \multirow{4}{*}{GT} & 0 & 14 & 1.0 & 1.0 & 1.0 & 1.0 & 1.0 &  & \tabularnewline
 &  & 29 & 38 & 0.921044 & 0.921044 & 0.921044 & 0.921044 & 0.921044 &  & \tabularnewline
 &  & 47 & 72 & 0.956282 & 0.955743 & 0.955743 & 0.955743 & 0.955743 &  & \tabularnewline
 &  & 104 & 125 & 0.985336 & 0.985336 & 0.985336 & 0.985336 & 0.985336 &  & \tabularnewline
\cmidrule(lr){2-11}
 & \multirow{4}{*}{KTH-SE} & 0 & 12 & 1.0 & 1.0 & 1.0 & 1.0 & 1.0 & 1.0 & \tabularnewline
 &  & 27 & 36 & 0.98783 & 0.98783 & 0.98783 & 0.98783 & 0.98783 & 0.98783 & \tabularnewline
 &  & 45 & 71 & 0.90019 & 0.90019 & 0.90019 & 0.90019 & 0.90019 & 0.90019 & \tabularnewline
 &  & 106 & 121 & 0.92464 & 0.92464 & 0.92464 & 0.92464 & 0.92464 & 0.92464 & \tabularnewline
\cmidrule(lr){2-11}
 & \multirow{4}{*}{TUG-AT} & 0 & 14 & 1.0 & 1.0 & 1.0 & 1.0 & 1.0 &  & 1.0\tabularnewline
 &  & 29 & 38 & 1.0 & 1.0 & 1.0 & 1.0 & 1.0 &  & 0.999322\tabularnewline
 &  & 47 & 72 & 0.955311 & 0.955311 & 0.955311 & 0.955311 & 0.955311 &  & 0.954555\tabularnewline
 &  & 105 & 125 & 0.979573 & 0.978914 & 0.978914 & 0.978914 & 0.978914 &  & 0.984449\tabularnewline
\midrule
\multicolumn{11}{l}{Dataset: Fluo-N2DH-GOWT1}\tabularnewline
\midrule
\multirow{10}{*}{01} & \multirow{4}{*}{GT} & 0 & 15 & 0.986659 & 0.98249 & 0.98249 & 0.98249 & 0.98249 &  & \tabularnewline
 &  & 21 & 45 & 0.995921 & 0.995921 & 0.995921 & 0.995921 & 0.995921 &  & \tabularnewline
 &  & 51 & 76 & 1.0 & 1.0 & 1.0 & 1.0 & 1.0 &  & \tabularnewline
 &  & 77 & 91 & 0.980701 & 0.980701 & 0.980701 & 0.980701 & 0.980701 &  & \tabularnewline
\cmidrule(lr){2-11}
 & \multirow{3}{*}{KTH-SE} & 27 & 46 & 0.91849 & 0.91849 & 0.91849 & 0.91849 & 0.91849 & 0.91849 & \tabularnewline
 &  & 47 & 76 & 0.980322 & 0.980322 & 0.980322 & 0.980322 & 0.980322 & 0.980322 & \tabularnewline
 &  & 77 & 91 & 0.980701 & 0.980701 & 0.980701 & 0.980701 & 0.980701 & 0.980701 & \tabularnewline
\cmidrule(lr){2-11}
 & \multirow{3}{*}{TUG-AT} & 0 & 10 & 0.996174 & 0.996174 & 0.996174 & 0.996174 & 0.996174 &  & 0.996174\tabularnewline
 &  & 21 & 49 & 0.994108 & 0.994108 & 0.994108 & 0.994108 & 0.994108 &  & 0.994108\tabularnewline
 &  & 52 & 75 & 0.995403 & 0.995403 & 0.995403 & 0.995403 & 0.995403 &  & 0.995663\tabularnewline
\midrule
\multirow{2}{*}{02} & \multirow{1}{*}{GT} & 62 & 74 & 0.997753 & 0.997753 & 0.997753 & 0.997753 & 0.997753 &  & \tabularnewline
\cmidrule(lr){2-11}
 & \multirow{1}{*}{KTH-SE} & 37 & 48 & 0.966894 & 0.966894 & 0.966894 & 0.966894 & 0.966894 & 0.96369 & \tabularnewline
\bottomrule
\end{tabular}